\documentclass{article}
\usepackage{PRIMEarxiv}
\usepackage{multirow}
\usepackage{graphicx} 
\usepackage{algorithm}
\usepackage{algorithmic}
\usepackage{amsmath}
\usepackage{amsfonts}
\usepackage{amssymb}
\usepackage{amsbsy}
\usepackage{bbm}
\usepackage{xcolor}
\usepackage{hyperref}

\newcommand{\rd}{\mathrm{d}}

\newcommand{\addeps}{\epsilon I_d+}

\usepackage[utf8]{inputenc} 
\usepackage[T1]{fontenc}    
\usepackage{hyperref}       
\usepackage{url}            
\usepackage{booktabs}       
\usepackage{amsfonts}       
\usepackage{nicefrac}       
\usepackage{microtype}      
\usepackage{lipsum}
\usepackage{fancyhdr}       
\usepackage{graphicx}       
\graphicspath{{media/}}     
\newtheorem{assumption}{Assumption}
\newtheorem{lemma}{Lemma}
\newtheorem{theorem}{Theorem}
\newtheorem{definition}{Definition}
\newtheorem{corollary}{Corollary}

\title{Reusing Historical Trajectories in Natural Policy Gradient via Importance Sampling: Convergence and Convergence Rate
}

\author{
  Yifan Lin\textsuperscript{*}\\
  School of Industrial and Systems Engineering\\
  Georgia Institute of Technology \\
  Atlanta\\
  \texttt{yuhaowang@gatech.edu} \\
  \And
 Yuhao Wang\textsuperscript{*}\\
  School of Industrial and Systems Engineering\\
  Georgia Institute of Technology \\
  Atlanta\\
  \texttt{skim@isye.gatech.edu}
   \And
  Enlu Zhou \\
  School of Industrial and Systems Engineering\\
  Georgia Institute of Technology \\
  Atlanta\\
  \texttt{enlu.zhou@isye.gatech.edu} \\
}

\begin{document}
\maketitle
\noindent\textsuperscript{*}These authors contributed equally.

\begin{abstract}
Reinforcement learning provides a mathematical framework for learning-based control, whose success largely depends on the amount of data it can utilize. The efficient utilization of historical trajectories obtained from previous policies is essential for expediting policy optimization. Empirical evidence has shown that policy gradient methods based on importance sampling work well. However, existing literature often neglect the interdependence between trajectories from different iterations, and the good empirical performance lacks a rigorous theoretical justification. In this paper, we study a variant of the natural policy gradient method with reusing historical trajectories via importance sampling. We show that the bias of the proposed estimator of the gradient is asymptotically negligible, the resultant algorithm is convergent, and reusing past trajectories helps improve the convergence rate. We further apply the proposed estimator to popular policy optimization algorithms such as trust region policy optimization. Our theoretical results are verified on classical benchmarks.
\end{abstract}


\section{Introduction}\label{sec:intro}
In challenging reinforcement learning tasks with large state and action spaces, policy optimization methods rank among the most effective approaches. It provides a way to directly optimize policies and handles complex and high-dimensional policy representations such as neural networks, all of which contribute to its popularity in the field. It usually works with parameterized policies and employs a policy gradient approach to search for the optimal solution (e.g. \cite{sutton1999policy}). The gradients can be estimated using various techniques, such as the REINFORCE algorithm (\cite{williams1992simple}) or actor-critic methods (e.g. \cite{konda1999actor}). These gradient estimation techniques provide a principled way to update the policy parameters based on the observed rewards and state-action trajectories.

The aforementioned on-policy gradient approach involves an iterative approach of gathering trajectories (or samples, these two terms are used interchangeably in the paper) by interacting with the environment, typically using the current-learned policy. This trajectory is then utilized to improve the policy. However, in many scenarios, conducting online interactions can be impractical. This can be due to the high cost of data collection (e.g., in robotics \cite{kahn2020badgr} or healthcare \cite{tang2021model}) or the potential dangers involved (e.g., in autonomous driving \cite{fang2022offline}). Additionally, even in situations where online interaction is feasible, there may be a preference for utilizing previously collected data to improve the gradient estimation, especially when online data are scarce (\cite{peshkin2002learning}). 

Reusing historical trajectories to accelerate the learning of the optimal policy is typically achieved by using the importance sampling technique, which could be traced back to \cite{rubinstein1990optimization}. In reinforcement learning, importance sampling is widely used for off-policy evaluation (e.g. \cite{hanna2019importance}) and policy optimization (e.g. \cite{metelli2018policy}). One significant limitation of the importance sampling approach in policy optimization is that it can suffer from high variance caused by the importance weights, particularly when the trajectory is long (see \cite{mandel2014offline}). This often occurs in the episode-based approach, where the importance weight is built on the product of likelihood ratio of state-action transitions in each episode (see \cite{baxter2001infinite}). On the contrary, step–based approaches (see \cite{lillicrap2015continuous}), derived from the Policy Gradient Theorem (see \cite{sutton1999policy}), can estimate the gradient by averaging over timesteps. For example, \cite{liu2018breaking} propose to apply importance sampling directly on the discounted state visitation distributions to avoid the exploding variance. Recently, \cite{metelli2020importance} propose a policy optimization via importance sampling approach that mixes online and offline optimization to efficiently exploit the information contained in the collected trajectories. \cite{zheng2022variance} propose a variance reduction based experience replay framework that selectively reuses the most relevant trajectories to improve policy gradient estimation.

Apart from reusing historical trajectories via importance sampling to accelerate the convergence of policy gradient algorithm, natural gradient (see \cite{amari1998natural}) has also been introduced to accelerate the convergence by considering the geometry of the policy parameter space (see \cite{kakade2001natural}). It is observed that the natural policy gradient algorithm often results in more stable updates that can prevent large policy swings and lead to smoother learning dynamics (see \cite{kakade2001natural}). Another benefit of natural policy gradient is its invariance to the parameterization of the policy, which allows for greater flexibility in designing the policy representation (see \cite{amari1998natural}).

It should be noted that most of the existing works in importance sampling-based policy optimization assume the importance sampling-based gradient estimator is unbiased (e.g. \cite{metelli2018policy, metelli2020importance, zheng2022variance}), and the convergence analysis is based on the unbiased gradient estimator. However, it is pointed out in \cite{eckman2018reusing}, \cite{eckman2018green} and \cite{liu2020simulation} that the importance sampling-based gradient estimator is biased in the iterative approach due to the dependence across iterations. Regarding the biased gradient estimator, \cite{liu2020simulation} study the asymptotic convergence of the stochastic gradient descent (SGD) method with reusing historical trajectories. 

In this paper, we propose to use importance sampling in the natural policy gradient algorithm, where importance sampling is used to estimate the gradient as well as the Fisher information matrix (FIM). We extend the convergence analysis of SGD in the context of simulation optimization (\cite{liu2020simulation}) to natural policy gradient in the context of reinforcement learning. We theoretically study a mini-batch natural policy gradient with reusing historical trajectories (RNPG) and show the asymptotic convergence of the proposed algorithm by the ordinary differential equation (ODE) approach. We show that the bias of the natural gradient estimator with historical trajectories is asymptotically negligible, and RNPG shares the same limit ODE as the vanilla natural policy gradient (VNPG), which only uses trajectories of the current iteration for FIM and gradient estimators. The asymptotic convergence rate is characterized by the stochastic differential equation (SDE) approach, and reusing past trajectories in the gradient estimator can improve the convergence rate by an order of $O(\frac{1}{K})$, where we reuse previous $K-1$ iterations' trajectories. Furthermore, with a constant step size, we find that RNPG induces smaller estimation error using a non-asymptotic analysis around the local optima. We also demonstrate that the proposed RNPG can be applied to other popular policy optimization algorithms such as trust region policy optimization (TRPO, \cite{schulman2015trust}).


Our main contributions are summarized as follows.
\begin{enumerate}
    \item We propose a variant of natural policy gradient algorithm (called RNPG), which reuses historical trajectories via importance sampling and accelerates the learning of the optimal policy.
    \item We provide a rigorous asymptotic convergence analysis of the proposed algorithm by an ODE approach. We further characterize the improved convergence rate by an SDE approach. 
    \item We empirically study the choice of different reuse size in the proposed algorithm and demonstrate the benefit of reusing historical trajectories on classical benchmarks. 
\end{enumerate}

The rest of the paper is organized as follows. Section~\ref{sec:algorithm} gives the problem formulation and presents the RNPG algorithm. Section~\ref{sec:convergence} analyzes the convergence behavior of RNPG by the ODE method. Section~\ref{sec:rate} characterizes the convergence rate of RNPG by the SDE approach. Section~\ref{sec:numerical} demonstrates the performance improvement of RNPG over VNPG on classical benchmarks. Section~\ref{sec:conclusion} concludes the paper.

\section{Problem Formulation and Algorithm Design}\label{sec:algorithm}
In this section, we first introduce the Markov decision process (MDP) and briefly review the natural policy gradient algorithm. This on-policy gradient approach involves an iterative approach of gathering experience by interacting with the environment, typically using the currently learned policy. However, in many scenarios, conducting online interactions can be impractical. Additionally, even in situations where online interaction is feasible, there may be a preference for utilizing previously collected data to improve the gradient estimation, especially when online data are scarce. We then propose to reuse historical trajectories in the natural policy gradient and present our main algorithm.

\subsection{Preliminaries: Markov Decision Process}\label{sec:MDP}
Consider an infinite-horizon MDP defined as $(\mathcal{S}, \mathcal{A}, \mathcal{P},$ $\mathcal{R}, \gamma, \rho_0)$, where $\mathcal{S}$ is the state space, $\mathcal{A}$ is the action space, $\mathcal{P}$ is the transition probability with $\mathcal{P}(s_{t+1}|s_t,a_t)$ denoting the probability of transitioning to state $s_{t+1}$ from state $s_t$ when action $a_t$ is taken, $\mathcal{R}$ is the reward function with $\mathcal{R}(s_t,a_t)$ denoting the cost at time stage $t$ when action $a_t$ is taken and state transitions from $s_t$, $\gamma \in (0,1)$ is the discount factor, $\rho_0$ is the probability distribution of the initial state, i.e., $s_0 \sim \rho_0$. 

Consider a stochastic parameterized policy $\pi_{\theta}: \mathcal{S} \to \Delta(\mathcal{A})$, defined as a function mapping from the state space to a probability simplex $\Delta(\cdot)$ over the action space, parameterized by $\theta \in \mathbb{R}^{d}$. For a particular probability (density) from this distribution we write $\pi_{\theta}(a|s)$. There are a large number of parameterized policy classes. For example, in the case of direct parameterization, the policies are parameterized by $\pi_{\theta}(a | s)=\theta_{s, a}$, where $\theta \in \Delta(\mathcal{A})^{|\mathcal{S}|}$ is within the probability simplex on the action space. In the case of softmax parameterization, 
\begin{equation*}
    \pi_{\theta}(a \mid s)=\frac{\exp \left(\theta_{s, a}\right)}{\sum_{a^{\prime} \in \mathcal{A}} \exp \left(\theta_{s, a^{\prime}}\right)}.
\end{equation*} 
The policies can also be parameterized by neural networks, where significant empirical successes have been achieved in many challenging applications, such as playing Go (see \cite{silver2016mastering}). The goal is  to find an  optimal policy $\pi_{\theta^{*}}$ that maximizes the expected total discounted return:
\begin{equation}  \label{objective}
\theta^* \in  \arg\max_{\theta\in\Theta} \eta(\pi_\theta) :=\mathbb{E}_{s_{0}, a_{0}, \ldots}\left[\sum_{t=0}^{\infty} \gamma^{t} \mathcal{R}\left(s_{t}, a_{t}\right)\right],
\end{equation}
where $s_{0} \sim \rho_{0}\left(s_{0}\right), a_{t} \sim \pi_\theta\left(a_{t} \mid s_{t}\right), s_{t+1} \sim \mathcal{P} \left(s_{t+1} \mid s_{t}, a_{t}\right)$, $\Theta$ is the feasible set of $\theta$, and $\eta(\pi_\theta)$ is the expected total discounted return of policy $\pi_\theta$. 

Denote by $d^{\pi_{\theta}}(s)$ the discounted state visitation distribution induced by the policy $\pi_{\theta}$, $d^{\pi_{\theta}}(s)= (1-\gamma)\sum_{t=0}^{\infty} \gamma^{t} \mathcal{P}\left(s_{t}=s \mid \pi_{\theta}\right)$. 
It is useful to define the discounted occupancy measure as $d^{\pi_\theta}(s,a)=d^{\pi_\theta}(s)\pi_{\theta}(a|s)$. Using the discounted occupancy measure, we can rewrite the expected total discounted return as $\eta(\pi_\theta)=\mathbb{E}_{(s, a) \sim d^{\pi_\theta}(s,a)}[\mathcal{R}(s, a)]$. We use the following standard definitions of the value function $V^{\pi_\theta}$, the state-action value function $Q^{\pi_\theta}$, and the advantage function $A^{\pi_\theta}$: $V^{\pi_\theta}\left(s_t\right)=\mathbb{E}_{a_{t}, s_{t+1}, \ldots}\left[\sum_{l=0}^{\infty} \gamma^{l} R \left(s_{t+l},a_{t+l}\right)\right]$, $Q^{\pi_\theta}\left(s_{t}, a_{t}\right)=\mathbb{E}_{s_{t+1}, a_{t+1}, \ldots}\left[\sum_{l=0}^{\infty} \gamma^{l} R\left(s_{t+l},a_{t+l}\right)\right]$, and $A^{\pi_\theta}(s, a)=Q^{\pi_\theta}(s, a)-V^{\pi_\theta}(s)$. 

\subsection{Preliminaries: Natural Policy Gradient}\label{sec:NPG}
In the policy gradient algorithm, at each iteration $n$, we can iteratively update the policy parameters by
\begin{align*}
\theta_{n+1}=\operatorname{Proj}_{\Theta}\left(\theta_{n}+\alpha_{n} \nabla \eta \left(\theta_{n}\right)\right),
\end{align*}
where $\alpha_n$ is the step size, $\operatorname{Proj}_{\Theta}(\theta)$ is a projection operator that projects the iterate of $\theta$ to the feasible parameter space $\Theta$, and $\nabla \eta\left(\theta_{n}\right)$ is the policy gradient. For ease of notations, we use parameter $\theta$ to indicate a parameterized policy $\pi_{\theta}$. The gradient is taken with respect to $\theta$ unless specified otherwise. The policy gradient (e.g. \cite{sutton1999policy}) is given by
\begin{align*}
    \nabla \eta(\theta) = \frac{1}{1-\gamma}\mathbb{E}_{(s,a) \sim d^{\pi_{\theta}}(s,a)} [A^{\pi_\theta}(s,a) \nabla \log \pi_{\theta}(a|s)].
\end{align*}

The steepest descent direction of $\eta (\theta)$ in the policy gradient is defined as the vector $d \theta$ that minimizes $\eta (\theta + d \theta)$ under the constraint that the squared length $||d\theta||^2$ is held to a small constant. This squared length is defined with respect to some positive-definite matrix $F(\theta)$ such that $||d\theta||^2=d\theta^T F(\theta) d\theta$. The steepest descent direction is then given by $F^{-1}(\theta)\nabla \eta (\theta)$ (see \cite{amari1998natural}). It can be seen that the policy gradient descent is a special case where $F(\theta)$ is the identity matrix, and the considered parameter space $\Theta$ is Euclidean. The natural policy gradient (NPG) algorithm (see \cite{kakade2001natural}) defines $F(\theta)$ to be the Fisher information matrix (FIM) induced by $\pi_{\theta}$, and performs natural gradient descent as follows:
\begin{align}\label{eq:fisher}
    \theta_{n+1} = \operatorname{Proj}_{\Theta}\left(\theta_{n}+\alpha_{n} F^{-1}(\theta_n) \nabla \eta\left(\theta_{n}\right)\right),
\end{align}
where $F(\theta)=\mathbb{E}_{(s,a) \sim d^{\pi_\theta}(s,a)} [\nabla \log \pi_{\theta}(a | s)\left(\nabla \log \pi_{\theta}(a | s)\right)^T]$. In practice, both the FIM and policy gradient are estimated by samples. Specifically, at each $n$-th iteration in stochastic natural policy gradient, the policy parameter is updated by
\begin{align*}
    \theta_{n+1} = \operatorname{Proj}_{\Theta}\left(\theta_{n}+\alpha_{n} \widetilde{F}^{-1}(\theta_n) \widetilde{\nabla \eta}\left(\theta_{n}\right)\right),
\end{align*}
where $\widetilde{F}(\theta_n)$ and $\widetilde{\nabla \eta}\left(\theta_{n}\right)$ are estimators for FIM and policy gradient, respectively.

\subsection{Natural Policy Gradient with Reusing Historical Trajectories}\label{sec:RNPG}
For ease of notations, we denote by $\xi_{n}^{i}=(s_n^i,a_n^i)$ the $i$-th state-action pair sampled from the discounted occupancy measure $d^{\pi_{\theta_n}}(s,a)$ at iteration $n$. We assume $\{\xi_n^i, i=1,\cdots,B\}$ are independent and identically distributed (i.i.d.) samples from the occupancy measure induced by the policy $\pi_{\theta_n}$. The i.i.d. samples can be generated in the following way. First, one generates a geometry random variable $T$ with success probability $1-\gamma$, that is, $\mathcal{P}( T = t) = \gamma^t (1-\gamma)$. Next, one generates one trajectory with length $T$ by sampling $s_0 \sim \rho_0,a_t\sim\pi_\theta(a_t|s_t)$ for $t\le T$, $s_{t} \sim \mathcal{P}(s_{t+1}|s_t,a_t)$ for $t\le T-1$. We then have the final state-action pair $\xi:= (s_T,a_T)$ follows the occupancy measure $d^{\pi_\theta}$. Indeed, 
\begin{align*}
    &\mathcal{P}\left((s_T,a_T)=(s,a)\right) \\
    = & \mathbb{E}\left[ \mathcal{P}\left((s_T,a_T)=(s,a)| T\right)\right] \\
    = & \sum_{t=0}^\infty \mathcal{P}(T=t) \mathcal{P}\left(s_t=s\right)\mathcal{P}(a_t=a|s)\\
    = & (1-\gamma)\pi_\theta(a|s)\sum_{t=0}^\infty \gamma^t\mathcal{P}\left(s_t=s\right)\\
    =& d^{\pi_\theta}(s,a).
\end{align*}
The independence can be then satisfied by generating independent trajectories with random lengths. However, it is worth noting in this way only the last sample is utilized and it requires re-starting the environment for each sample. While the i.i.d. assumption is necessary to demonstrate the convergence of the proposed algorithm, as also assumed in, e.g., \cite{qiu2021finite, kumar2023sample}, in practice, the algorithm is usually implemented in a more sample-efficient way, such as single-path generation in \cite{schulman2015trust}. The empirical efficiency of the proposed algorithm is demonstrated in Section \ref{sec:numerical}, even when i.i.d. data are not available.
A vanilla baseline gradient estimator $\widetilde{\nabla \eta}\left(\theta_{n}\right)$ and  FIM estimator $\widetilde{F}\left(\theta_{n}\right)$ can be obtained as:
\begin{align*}
   \widetilde{\nabla \eta}\left(\theta_{n}\right) = \frac{1}{B} \sum_{i=1}^{B} G(\xi_n^i,\theta_n), \quad     \widetilde{F}\left(\theta_{n}\right) = \frac{1}{B} \sum_{i=1}^{B} S(\xi_n^i,\theta_n),
\end{align*}
where $S(\xi,\theta)=\nabla \log \pi_{\theta}(a|s) (\nabla \log \pi_{\theta}(a|s))^{T}$ and $G(\xi,\theta)=\frac{1}{1-\gamma}A^{\pi_{\theta}}(s,a)\nabla \log \pi_{\theta}(a|s)$, 
where $\xi$ appears in $A^{\pi_\theta}(s,a)$ and $\pi_{\theta}(a|s)$. It is easy to see that $\widetilde{F}\left(\theta_{n}\right)$ and $\widetilde{\nabla \eta}\left(\theta_{n}\right)$ are unbiased estimators of the FIM $F(\theta_n)$ and the gradient $\nabla \eta\left(\theta_{n}\right)$, respectively. However, in the vanilla stochastic natural policy gradient (VNPG), a small batch size $B$, which is often the case when there is limited online interaction with the environment, could lead to a large variance in the estimator. An alternative gradient and FIM estimator, which reuse historical trajectories, are as follows:
\begin{align}\label{eq:gradient_reuse}
    \widehat{\nabla \eta}(\theta_n)=\frac{1}{K_1B} \sum_{m=n-K_1+1}^{n} \sum_{i=1}^{B} \omega(\xi_m^i, \theta_n | \theta_m) G(\xi_m^i, \theta_n),
\end{align}
\begin{align}
    \widehat{F}\left(\theta_{n}\right) = \frac{1}{K_2B} \sum_{m=n-K_2+1}^{n} \sum_{i=1}^{B} \omega(\xi_m^i, \theta_n | \theta_m) S(\xi_m^i,\theta_n), \notag
\end{align}

where we reuse previous $K_1-1$ iterations' trajectories for estimating the gradient and $K_2-1$ iterations' trajectories for estimating FIM. The likelihood ratio $\omega(\xi_m^i, \theta_n | \theta_m)$ is given by
\begin{align}\label{eq:likelihood_ratio}
    \omega(\xi_m^i, \theta_n | \theta_m)=\frac{d^{\pi_{\theta_n}}(\xi_m^i)}{d^{\pi_{\theta_m}}(\xi_m^i)}.
\end{align}
Moreover, since we need to take the inverse of $\widehat{F}^{-1}(\theta_n)$, for numerical stability, we add a regularization term $\epsilon I_d$ to $\widehat{F}(\theta_n)$ to make it strictly positive definite, where $\epsilon>0$ is a small positive number and $I_d$ is a $d$-by-$d$ identity matrix. Therefore, 
\begin{align}\label{eq:FIM_reuse}
    \widehat{F}\left(\theta_{n}\right) = \epsilon I_d + \frac{1}{K_2B} \sum_{m=n-K_2+1}^{n} \sum_{i=1}^{B} \omega(\xi_m^i, \theta_n | \theta_m) S(\xi_m^i,\theta_n).
\end{align}
The update of the natural policy gradient with reusing historical trajectories (RNPG) is then as follows. 
\begin{align}\label{eq:update_reuse}
    \theta_{n+1} = \operatorname{Proj}_{\Theta}\left(\theta_{n}+\alpha_{n} \widehat{F}^{-1}(\theta_n)  \widehat{\nabla \eta}\left(\theta_{n}\right)\right).
\end{align}

We summarize RNPG in Algorithm 1.

\textbf{Algorithm 1: Natural Gradient Descent with Reusing Historical Trajectories} \vspace{-2mm}
\begin{itemize}
    \item[1.] At iteration $n=0$, choose an initial parameter $\theta_0$. Draw i.i.d. samples $\{\xi_0^i,i=1,\cdots,B\}$ from discounted occupancy measure $d^{\pi_{\theta_0}}(s,a)$ by interacting with the environment.
    \item[2.] At iteration $n + 1$, conduct the following steps.
    \begin{itemize}
        \item[2.1] Update $\theta_{n+1}$ according to \eqref{eq:update_reuse}.
        \item[2.2] Draw i.i.d. samples $\{\xi_{n+1}^i,i=1,\cdots,B\}$ from discounted occupancy measure $d^{\pi_{\theta_{n+1}}}(s,a)$ by interacting with the environment. 
        \item[2.3] $n = n + 1$. Repeat the procedure 2. 
    \end{itemize}
    \item[3.] Output $\theta_{n}$ and $\pi_{\theta_n}$ when some stopping criteria are satisfied.
\end{itemize}

From a computational perspective, it should also be noted that the likelihood ratio in \eqref{eq:FIM_reuse} and \eqref{eq:gradient_reuse} is usually hard to compute, since the discounted occupancy measure does not admit a closed form expression. We defer the discussion to Section~\ref{sec:convergence} on some approximations to make Algorithm 1 more practical.


\subsection{Bias and Variance of the Gradient Estimator Reusing Samples from Previous Iteration}

Some prior work, including  \cite{liu2020simulation},  \cite{eckman2018reusing}, and \cite{eckman2018green}, have considered reusing past samples via the likelihood ratio approach in simulation optimization problems and noted that the dependence between iterations introduces bias into their gradient estimators. Similar to their observations, such iteration dependence also introduces bias into our FIM and gradient estimators when reusing historical trajectories. 
Let's use a two-iteration example ($K_1 = K_2 = K = 2$) to illustrate this bias. Consider the first step of RNPG given a deterministic initial solution $\theta_0 \in \Theta$. Note that both the FIM estimator $\widehat{F}(\theta_0)$ and gradient estimator $\widehat{\nabla \eta}(\theta_0)$ are based on $B$ replications run at $\theta_0$. Therefore, we have $\theta_1=\operatorname{Proj}_{\Theta}\left(\theta_0+\alpha_0(\widehat{F}^{-1}(\theta_0) \widehat{\nabla \eta}(\theta_0))\right)$. For simplicity, assume no projection is needed for $\theta_1$, thus $\theta_1=\theta_0+\alpha_0(\widehat{F}^{-1}(\theta_0) \widehat{\nabla \eta}(\theta_0))$. Also note that $\widehat{\nabla \eta}(\theta_1)=\frac{1}{2B}\sum_{i=1}^{B}G(\xi_0^i, \theta_1)\omega(\xi_0^i,\theta_1|\theta_0) + \frac{1}{2B}  \sum_{i=1}^{B}G(\xi_1^i, \theta_1)$, $\widehat{F}(\theta_1)=\frac{1}{2}\left(\epsilon I_d + \frac{1}{B}\sum_{i=1}^{B}S(\xi_0^i, \theta_1)\omega(\xi_0^i,\theta_1|\theta_0)\right) + \frac{1}{2}  \left(\epsilon I_d + \frac{1}{B}\sum_{i=1}^{B}S(\xi_1^i, \theta_1)\right)$. Then the expectation of the gradient estimator in RNPG is 
\begin{align*}
    \mathbb{E}[\widehat{\nabla \eta}(\theta_1)|\theta_1] = \underbrace{\frac{1}{2} \mathbb{E}\left[\frac{1}{B}\sum_{i=1}^{B}G(\xi_0^i,\theta_1)\omega(\xi_0^i,\theta_1|\theta_0) \bigg| \theta_1\right]}_{Z} + \frac{1}{2}\nabla \eta(\theta_1).
\end{align*}
Since $\theta_1 = \theta_0+\alpha_0\left(\epsilon I_d+\frac{1}{B}\sum_{i=1}^{B}S(\xi_0^i,\theta_0)\right)^{-1}\frac{1}{B}\sum_{i=1}^{B}G(\xi_0^i,\theta_0)$ depends on $\xi_0^i, i=1,2,\ldots,B$, $\xi_0^i | \theta_1$ does not follow the occupancy measure $d^{\pi_{\theta_0}}$. This means
\begin{align*}
  &\mathbb{E}\left[ G(\xi_0^i,\theta_1)\omega(\xi_0^i,\theta_1|\theta_0) \bigg| \theta_1\right] \\
  \neq & \int_{\xi}  G(\xi,\theta_1)\omega(\xi,\theta_1|\theta_0) d^{\pi_{\theta_0}}(\xi) \mathrm{d}\xi\\
  = & \int_{\xi}  G(\xi,\theta_1)\frac{d^{\pi_{\theta_1}}(\xi)}{d^{\pi_{\theta_0}}(\xi)} d^{\pi_{\theta_0}}(\xi) \mathrm{d}\xi\\
  = & \int_{\xi}  G(\xi,\theta_1)d^{\pi_{\theta_1}}(\xi) \mathrm{d}\xi\\
  =& \nabla\eta(\theta_1).
\end{align*}
Hence, $Z = \frac{1}{2}\sum_{i=1}^B\mathbb{E}\left[ G(\xi_0^i,\theta_1)\omega(\xi_0^i,\theta_1|\theta_0) \bigg| \theta_1\right]\neq \nabla \eta (\theta_1) $.

By extension, the natural policy gradient estimator in RNPG is also biased.
In summary, the conditional distribution of $\xi_0^i$ given $\theta_1$ differs from the distribution from which $\xi_0^i$ was originally sampled.

In addition to the bias, let's also consider the variance of the natural gradient estimator when using historical trajectories. Similarly, consider the gradient estimator using samples from the first and second iteration as discussed above. We can write the variance of the gradient estimator (conditioned on $\theta_1$) as:
\begin{align} \label{eq: variance first iter}
    \operatorname{VaR}\left(\widehat{\nabla \eta}(\theta_1)\right) =  \underbrace{\frac{1}{4} \operatorname{VaR}\left(\frac{1}{B}\sum_{i=1}^{B}G(\xi_0^i,\theta_1)\omega(\xi_0^i,\theta_1|\theta_0) \bigg| \theta_1\right)}_{Z'} + \frac{1}{4B} \operatorname{VaR}(G(\xi,\theta_1)),
\end{align}
where the covariance term 
\begin{align*}
    &2\mathbb{E}\left[ \left( \frac{1}{2B}\sum_{i=1}^{B}G(\xi_0^i,\theta_1)\omega(\xi_0^i,\theta_1|\theta_0) - Z\right)\left( \frac{1}{B}\sum_{i=1}^{2B}G(\xi_1^i,\theta_1) - \frac{1}{2}\nabla \eta(\theta_1)\right) \bigg|\theta_1\right]\\
    =& 2\mathbb{E}\left[ \left( \frac{1}{2B}\sum_{i=1}^{B}G(\xi_0^i,\theta_1)\omega(\xi_0^i,\theta_1|\theta_0) - Z\right)\bigg|\theta_1\right]\mathbb{E}\left[\left( \frac{1}{B}\sum_{i=1}^{2B}G(\xi_1^i,\theta_1) - \frac{1}{2}\nabla \eta(\theta_1)\right) \bigg|\theta_1\right]\\
    =&0,
\end{align*}
where the first equality holds since conditioned on $\theta_1$, $\xi_1^i$ is independent of past samples. From \eqref{eq: variance first iter}, compared with the variance of the gradient estimator only using the current samples, which is $\frac{1}{B}\operatorname{VaR}(G(\xi,\theta_1))$, reusing samples in 
$\widehat{\nabla \eta}(\theta_1)$ will reduce the conditional variance if $Z' \le \frac{3}{4B} \operatorname{VaR}\left(G(\xi,\theta_1)\right)$  and increase the conditional variance otherwise.  The value of $Z'$  depends on the specific form of the importance ratio $\omega$ and the random gradient $G$.

Nonetheless, from an asymptotic perspective, reusing past samples is beneficial. Before diving into the rigorous  analysis, we provide some intuitive explanations here. First, as discussed above, the bias introduced by reusing samples is due to the dependence of current solution on past samples. With a diminishing step size sequence $\{\alpha_n\}$, the difference between solutions $\theta_n$ and $\theta_{n+\ell}$ with a fixed $\ell$ vanishes as $n$ goes to infinity. This indicates $\theta_{n+\ell}\approx \theta_n$ and $\omega(\cdot,\theta_{n+\ell}|\theta_n)\approx 1$ when $n$ is large, and as a result, asymptotically, we can view all samples 
generated  between iterations $n$ and $n+\ell$ as approximately conditional i.i.d. samples under parameter $\theta_{n+\ell} \approx \theta_n$. This further indicates
the bias caused by the difference among distributions of samples from iteration $n$ to $n+\ell$ also decreases to $0$ asymptotically. Similarly for the variance, as reused samples become asymptotically i.i.d., we expect the conditional variance of the gradient estimator decreases as the reuse size $K$ increases.
 In summary, we expect by reusing samples, the bias is negligible and the variance is reduced asymptotically. Throughout the remaining paper, we will show this rigorously and demonstrate numerically.

\subsection{Summary of Main Theoretical Results}

 We first give an intuitive summary of the main theoretical results in this paper, and an interested reader can go on to the next two sections for the detailed analysis. Choosing $K_1 = K, K_2 = 1$ (i.e., using samples of the most recent $K$ iterations in the gradient estimator, and only using samples of the current iteration in the FIM estimator), we show the following results:

\begin{itemize}
    \item[(1)] the solution trajectory $\{\theta_n\}$ given by RNPG (Algorithm 1) converges with probability one (w.p.1) to a local optimum $\bar{\theta}$, which is the same as the vanilla natural policy gradient (VNPG).  
    \item[(2)]the random error $\theta_n - \Bar{\theta}$  follows a normal distribution  $ \mathcal{N}(0,\alpha_n \Sigma_\infty)$ asymptotically, where $\alpha_n$ is the step size and $\Sigma_\infty \propto \frac{1}{B}\Sigma_1(\Bar{\theta}) +\frac{1}{KB} \Sigma_2(\Bar{\theta})$. The definition of $\Sigma_\infty$, $\Sigma_1$ and  $\Sigma_2$ will be explained in Section~\ref{sec:rate}. In essence,  $\sqrt{\alpha_n}A$, where $A$ satisfies $A A^T = \Sigma_\infty$,  characterizes the rate of convergence of $\theta_n$. We show that the proposed RNPG algorithm improves the convergence rate via decreasing the magnitude of $\Sigma_2$ by a factor of $\frac{1}{K}$.
\end{itemize}

\section{Convergence Analysis}\label{sec:convergence}
In this section, we analyze the convergence behavior of RNPG by the ordinary differential equation (ODE) method. Throughout the rest of the paper, we only reuse historical trajectories on the gradient estimator $\widehat{\nabla \eta}(\theta_n)$, while we use trajectories generated by the current policy to estimate the inverse FIM $\widehat{F}^{-1}(\theta_n) = \widetilde{F}^{-1}(\theta_n) = \left(\epsilon I_d+\frac{1}{B} \sum_{i=1}^{B} S(\xi_n^i,\theta_n)\right)^{-1}$. 
Not reusing samples for the inverse FIM helps simplify the analysis, as correlated samples in the inverse FIM estimator are extremely challenging to handle due to the matrix inversion. Also, our numerical results show that reusing historical trajectories in the inverse FIM estimator,  compared to reusing in the gradient estimator, yields only a minor performance improvement. For clarity, we denote the reuse size for the gradient estimator by \( K \) instead of \( K_1 \).


We show that RNPG and VNPG share the same limit ODE, while the bias resulting from the interdependence between iterations gradually diminishes, ultimately becoming insignificant in the asymptotic sense. We then propose some approximations to make the proposed algorithm more practical, and apply the proposed algorithm to some popular policy optimization algorithms such as TRPO.

\subsection{Regularity Conditions for RNPG}\label{sec:assumption}
We study the asymptotic behavior of Algorithm~1 by the ODE method (please refer to \cite{Kushner2003} for a detailed exposition on the ODE method for stochastic approximation). The main idea is that stochastic gradient descent (SGD, and in our case is stochastic natural policy gradient, NPG) can be viewed as a noisy discretization of an ODE. Under certain conditions, the noise in NPG averages out asymptotically, such that the NPG iterates converge to the solution trajectory of the ODE. 

We first summarize the regularity conditions for RNPG that are used throughout the paper. For any $s > 0$, let $\pmb{\xi}_s:=(\xi_s^1,\cdots,\xi_s^{B})$, $\pmb{d}_s:=(d^{\pi_{\theta}}(\xi_s^1),\cdots,d^{\pi_{\theta}}(\xi_s^B))$, effective memory $\pmb{e}_s:=(\pmb{\xi}_{s-K+1}$, $\pmb{d}_{s-K+1}, \cdots, \pmb{\xi}_{s-1}, \pmb{d}_{s-1})$, and non-decreasing filtration $\mathcal{F}_n:=\sigma\{(\theta_s, \pmb{e}_s), s \leq n\}$.

\begin{assumption}~
\begin{itemize}
    \item (A.1.1) The step size sequence $\{\alpha_n\}$ satisfies $\sum_{n=0}^{\infty} \alpha_{n}^{2}<\infty$, $\sum_{n=0}^{\infty} \alpha_{n}=\infty$, $\lim_{n \to \infty} \alpha_{n} = 0$, $\alpha_n > 0, \forall n \geq 0$. 
    \item (A.1.2) The absolute value of the reward $\mathcal{R}(s, a)$ is bounded uniformly, i.e., $\forall (s,a) \in \mathcal{S} \times \mathcal{A}$, there exists a constant $U_r>0$ such that $|\mathcal{R}(s, a)| \leq U_r$. 
    \item (A.1.3) The policy $\pi_{\theta}$ is differentiable with respect to $\theta$, Lipschitz continuous in $\theta$, and has strictly positive and bounded norm uniformly. That is, there exist constants $L_{\Theta}, U_{\Theta}, $ such that $\left\|\nabla \pi_{\theta_1}(a|s)-\nabla \pi_{\theta_2}(a|s)\right\| \leq L_{\Theta}\left\|\theta_1-\theta_2\right\|$, $\forall \theta_1, \theta_2 \in \Theta$, $\left\|\nabla \pi_{\theta}(a|s)\right\| \leq U_{\Theta}$, $\forall (s,a) \in \mathcal{S} \times \mathcal{A}$.
    \item (A.1.4) $\left\|\pi_{\theta_1}(\cdot|s)-\pi_{\theta_2}(\cdot|s)\right\|_{TV} \leq U_{\Pi}\left\|\theta_1-\theta_2\right\|, \forall \theta_1,\theta_2 \in \Theta, \forall s \in \mathcal{S}$, for some constant $U_{\Pi}>0$,
    where $\left\|P-Q\right\|_{TV}$ stands for total variation norm between two probability distributions $P$ and $Q$ with support $x$, i.e., $\left\|P-Q\right\|_{TV}=\frac{1}{2}\int_{x} |P(x)-Q(x)|dx$.
    \item (A.1.5) There exists a constant $\epsilon_d>0$ such that the discounted occupancy distribution $d^{\pi_{\theta}}(s,a) \geq \epsilon_d, \forall (s,a) \in \mathcal{S} \times \mathcal{A}, \forall \theta \in \Theta$.  
    \item (A.1.6) $\Theta$ is a nonempty compact and convex set in $\mathbb{R}^d$.
    \item (A.1.7) The FIM estimator $\widehat{F}(\theta_n)$ and gradient estimator $\widehat{\nabla \eta}(\theta_n)$ are conditionally independent given $\pmb{e}_n$.
\end{itemize}
\label{ass:1}
\end{assumption}

(A.1.1) essentially requires the step size diminishes to zero not too slowly ($\sum_{n=0}^{\infty} \alpha_{n}^{2}<\infty$) nor too quickly ($\sum_{n=0}^{\infty} \alpha_{n}=\infty$). For example, we can choose $\alpha_n=\frac{\alpha}{n}$ for some $\alpha>0$. (A.1.2) and (A.1.3) are standard assumptions on the regularity of the MDP problem and the parameterized policy. 
(A.1.4) holds for any smooth policy with bounded action space (see, e.g. \cite{xu2020improving}). (A.1.5) ensures the discounted occupancy distribution is bounded away from zero to ensure computational stability. This assumption implies that the state and action space is bounded, which is a general assumption (e.g., \cite{van2012reinforcement,schulman2015trust}).
(A.1.6) guarantees the uniqueness of the projection in the solution iterate. (A.1.7) is easily satisfied if we use independent samples for the FIM estimator and gradient estimator, respectively. For example, in each iteration, we could use $B$ i.i.d. samples for the FIM estimator and another $B$ i.i.d. samples for the gradient estimator. 

\subsection{Asymptotic Convergence by the ODE Method}\label{sec:ODE}
Before proceeding to our main convergence result, we introduce the continuous-time interpolation of the solution sequence $\{\theta_n\}$. Define $t_0=0$ and $t_n=\sum_{i=0}^{n-1}\alpha_i, n \geq 1$. For $t \geq 1$, let $N(t)$ be the unique $n$ such that $t_n \leq t < t_{n+1}$. For $t < 0$, set $N(t)=0$. Define the interpolated continuous process $\theta^{0}$ as $\theta^{0}(0) = \theta_0$ and $\theta^{0}(t) = \theta_{N(t)}$ for any $t > 0$, and the shifted process as $\theta^{n}(s) = \theta^{0}(s + t_n)$. 
We then show in the following theorem the limiting behavior of the solution trajectory in Algorithm~1.

\begin{theorem}
Let $\mathcal{D}^d[0,\infty)$ be the space of $\mathbb{R}^d$-valued operators which are right continuous and have left-hand limits for each dimension. Under Assumption~\ref{ass:1}, there exists a process $\theta^*(t)$ to which some subsequence of $\{\theta^n(t)\}$ converges w.p.1 in the space $\mathcal{D}^d[0,\infty)$, where $\theta^*(t)$ satisfies the following ODE
\begin{align}\label{eq:ODE}
\dot{\theta}= \Bar{F}^{-1}(\theta) \nabla \eta(\theta)+z, ~z \in -\mathcal{C}(\theta),
\end{align}
where $\Bar{F}^{-1}(\theta) = \mathbb{E}\left[\left(\addeps \frac{1}{B}\sum_{i=1}^B S(\xi_i,\theta) \right)^{-1} \right]$, $\xi_1,\ldots,\xi_B$ are $i.i.d.$ samples from the occupancy measure $d^{\pi_{\theta}}$ and $\mathcal{C}(\theta)$ is the Clarke's normal cone to $\Theta$, $z$ is the projection term, which is the maximum force needed to keep the trajectory of the ODE \eqref{eq:ODE} from leaving the parameter space $\Theta$. The solution trajectory $\{\theta_n\}$ in Algorithm~1 also converges w.p.1 to the limit set of the ODE \eqref{eq:ODE}.
\label{thm:ODE}
\end{theorem}

When any solution on the boundary of $\Theta$ is not a local maximum, Theorem~\ref{thm:ODE} indicates that the solution trajectory $\{\theta_n\}$ in Algorithm~1 satisfies $\nabla \eta(\theta_n) \rightarrow 0$ w.p.1.. However, this does not imply the global optimality. Whether the algorithm can achieving global optimality depends on the policy parameterization and the specific problem at hand. For instance, in a simple tabular MDP with direct policy parameterization (i.e., \(\pi(\cdot|s)\) for all \(s \in \mathcal{S}\)), any local optimum is also a global optimum, so NPG with sample reuse will converge to this global optimum. However, in more general MDPs with continuous state or action spaces and more complex policy parameterizations, neither NPG with nor without sample reuse is guaranteed to reach the global optimum. In practice, neural networks are often used to parameterize policies due to their powerful function approximation abilities, which allows for a more flexible policy representation but sacrifices the theoretical guarantee of global optimality.

Before the formal proof of Theorem~\ref{thm:ODE}, we first give a high-level proof outline. Note that in the update \eqref{eq:update_reuse}, we can decompose the natural gradient estimation into three components: the true natural gradient, the noise caused by the simulation error, and the bias caused by reusing historical trajectories. We then separately analyze the noise and bias effects on the estimation of FIM and gradient, and show the noise and bias terms are asymptotically negligible. 

With an explicit projection term $z_n$, we can rewrite \eqref{eq:update_reuse} as follows
\begin{align}\label{eq:decompose}
    \theta_{n+1} = \theta_{n} + \alpha_n\Big(\Bar{F}^{-1}(\theta_n)\nabla \eta(\theta_n) & + \underbrace{\Bar{F}^{-1}(\theta_n)\widehat{\nabla \eta}(\theta_n) - \Bar{F}^{-1}(\theta_n) \mathbb{E}[ \widehat{\nabla \eta}(\theta_n)|\mathcal{F}_n]}_{\delta M_n} \nonumber \\
    & + \underbrace{\bar{F}^{-1}(\theta_n)\mathbb{E}[ \widehat{\nabla \eta}(\theta_n)|\mathcal{F}_n] - \bar{F}^{-1}(\theta_n)\nabla \eta(\theta_n)}_{\zeta_n} \nonumber \\
    & + \underbrace{(\widehat{F}^{-1}(\theta_n)-\bar{F}^{-1}(\theta_n)) \widehat{\nabla \eta}(\theta_n)}_{\delta F_n} + z_n\Big),
\end{align}
where $\delta M_n$ is the noise term caused by the simulation error in the gradient estimator, $\zeta_n$ is the bias term caused by reusing historical trajectories in gradient estimator, and $\delta F_n$ is due to the simulation error in the inverse FIM estimator.
We will then show in the rest of the section that the continuous-time interpolations of $\delta M_n$, $\zeta_n$ and $\delta F_n$ do not change asymptotically. The formal definition of zero asymptotic rate of change is given below, which is from Chapter 5.3 in \cite{Kushner2003}. 

\begin{definition}[Zero asymptotic rate of change] A stochastic process $X(t)$ is said to have zero asymptotic rate of change w.p.1 if for some positive number $T$, 
\begin{align*}
    \lim_{n} \sup_{j \geq n} \max_{0 \leq t \leq T}\left|X(jT+t)-X(jT)\right|=0 \text { w.p.1}.
\end{align*}
\end{definition}

We first have the following lemma to show the continuous-time interpolations of the noise terms $\delta M_n$ and $\delta F_n$ have zero asymptotic rate of change. 

\begin{lemma}\label{lemma:noise}
Let the continuous-time interpolations of $\delta M_n$ and $\delta F_n$ be $M(t)=\sum_{i=0}^{N(t)-1} \alpha_i \delta M_i$ and $H(t)=\sum_{i=0}^{N(t)-1} \alpha_i \delta F_i$, respectively. Then $M(t)$ and $H(t)$ have zero asymptotic rate of change w.p.1 under Assumption~\ref{ass:1}.
\end{lemma}

We then adopt the fixed-state method from \cite{Kushner2003} to show the continuous-time interpolation of the bias term $\zeta_n$ has zero asymptotic rate of change. Let $P(\pmb{e}_{n+1}|\pmb{e}_n, \theta_n)$ be the transition probability given the current iterate $\theta_n$. Note that $\pmb{e}_n=(\pmb{\xi}_{n-K+1},\pmb{d}_{n-K+1}$,
$\cdots,\pmb{\xi}_{n-1},\pmb{d}_{n-1})$, $\pmb{e}_{n+1}=(\pmb{\xi}_{n-K+2},\pmb{d}_{n-K+2},\cdots,\pmb{\xi}_{n},\pmb{d}_{n})$. Given $\pmb{e}_n$, the component of $\pmb{e}_{n+1}$ that remains unknown are $\pmb{\xi}_n$ and $\pmb{d}_n$, which are random variables that only depend on $\theta_n$. Then $\pmb{e}_n$ has the Markov property: $P(\pmb{e}_{n+1}|\pmb{e}_m, \theta_m, m \leq n)=P(\pmb{e}_{n+1}|\pmb{e}_n, \theta_n)$. For a fixed state $\theta$, the transition probability $P(\pmb{e}'|\pmb{e}, \theta)$ defines a Markov chain denoted as $\{\pmb{e}_n(\theta)\}$. We expect that the probability law of the chain for a given $\theta$ is close to the probability law of the true $\{\pmb{e}_n\}$ if $\theta_n$ varies slowly around $\theta$. We are interested in $\{\pmb{e}_i(\theta_n):i \geq n\}$ with initial condition $\pmb{e}_n(\theta_n)=\pmb{e}_n$. Thus, this process starts at value $\pmb{e}_n$ at time $n$ and evolves as if the parameter value were fixed at $\theta_n$ forever after, and the limit ODE obtained in terms of this fixed-state chain approximates that of the original iterates. 

To explicitly express the estimators' dependence on the history of data $\pmb{e}_m$, $m \geq K - 1$, let 
\begin{align*}
    \widehat{\nabla \eta}(\theta, \pmb{e}_m)=\frac{1}{KB}\sum_{j=m-K+1}^{m}\sum_{i=1}^{B}\frac{d^{\pi_\theta}(\xi_j^i)}{d^{\pi_{\theta_j}}(\xi_j^i)}G(\xi_j^i,\theta).
\end{align*}

It is easy to check $\widehat{\nabla \eta}(\theta_n, \pmb{e}_n)=\widehat{\nabla \eta}(\theta_n)$. Define the function $v_n(\theta, \pmb{e}_n)$ as follows:
\begin{align*}
    v_n(\theta, \pmb{e}_n) & = \sum_{i=n}^{\infty} \alpha_i \bar{F}^{-1}(\theta)\mathbb{E}[\widehat{\nabla \eta}(\theta, \pmb{e}_i(\theta)) - \nabla \eta(\theta)|\pmb{e}_n(\theta)=\pmb{e}_n, \theta].
\end{align*}

$v_n(\theta, \pmb{e}_n)$ represents the accumulated bias brought by reusing historical trajectories in the gradient estimator, in the fixed-state chain with fixed state $\theta$. Next we show the bias in the fixed-state chain with fixed state $\theta_n$ vanishes. 

\begin{lemma}
Under Assumption~\ref{ass:1}, $\lim_{n \to \infty} v_n(\theta_n, \pmb{e}_n)=0$ w.p.1.
\label{lemma:bias}
\end{lemma}

We then consider a perturbed iteration $\widetilde{\theta}_n=\theta_n - v_n(\theta_n,\pmb{e}_n)$. 
For the gradient estimator, an error $b_n$ (due to the replacement of $\theta_{n+1}$ by $\theta_n$ in $v_{n+1}(\theta_{n+1},\pmb{e}_{n+1})$) and a new martingale difference term $\delta B_n$ were introduced in the process. We refer the readers to Chapter 6.6 in \cite{Kushner2003} for the detailed discussion on the perturbation. Lemma~\ref{lemma:bias} implies the perturbed iteration $\widetilde{\theta}_n$ asymptotically equals to $\theta_n$. We can rewrite the perturbed iteration as follows:
\begin{align*}
    \widetilde{\theta}_{n+1}=\widetilde{\theta}_{n} + \alpha_n \big(\bar{F}^{-1}(\theta_n)\nabla \eta(\theta_n) + \delta M_n + \delta F_n + z_n\big)+ b_n + \delta B_n,
\end{align*}
where $b_n=v_{n+1}(\theta_{n+1},\pmb{e}_{n+1}) - v_{n+1}(\theta_{n},\pmb{e}_{n+1})$, $\delta B_n = v_{n+1}(\theta_{n},\pmb{e}_{n+1}) - \mathbb{E}[v_{n+1}(\theta_{n},\pmb{e}_{n+1})|\pmb{e}_n(\theta)=\pmb{e}_n,\theta_n]$. Our next step is to show the continuous-time interpolations of $b_n$ and $\delta B_n$ have zero asymptotic rate of change. 

\begin{lemma}
Let the continuous-time interpolations of $b_n$ and $\delta B_n$ be $B(t)=\sum_{i=0}^{N(t)-1} b_i$ and $I(t)=\sum_{i=0}^{N(t)-1} \delta B_i$, respectively. Then $B(t)$, $I(t)$ have zero asymptotic rate of change w.p.1 under Assumption~\ref{ass:1}.
\label{lemma:fixed_chain}
\end{lemma}

We can then relate the bias term $\zeta_n$ in \eqref{eq:decompose} to $b_n$ and $\delta B_n$, and show the corresponding continuous-time interpolations have zero asymptotic rate of change in the next corollary.

\begin{corollary}\label{corollary:bias}
Let the continuous-time interpolation of $\zeta_n$ be $Z(t)=\sum_{i=0}^{N(t)-1} \alpha_i \zeta_i$. Then $Z(t)$ has zero asymptotic rate of change w.p.1 under Assumption~\ref{ass:1}.
\end{corollary}

We are now ready to show the formal proof of Theorem~\ref{thm:ODE}.

\textbf{Proof of Theorem~\ref{thm:ODE}}
The update \eqref{eq:update_reuse} in Algorithm~1 can be written as:
\begin{align*}
    \theta_{n+1} = \theta_n + \alpha_n \big(\bar{F}^{-1}(\theta_n)\nabla \eta(\theta_n) + \delta M_n + \zeta_n + \delta F_n + z_n\big),
\end{align*}
where $\delta M_n$ is the noise term caused by the simulation error in the gradient estimator, $\zeta_n$ is the bias term caused by reusing historical trajectories in gradient estimator, $\delta F_n$ is the noise term caused by the simulation error in the FIM estimator. By Lemma~\ref{lemma:noise}, the continuous-time interpolations of $\delta M_n$ and $\delta F_n$ have zero asymptotic rate of change. By Corollary~\ref{corollary:bias}, the continuous-time interpolation of $\zeta_n$ has zero asymptotic rate of change. Therefore, the limit ODE is determined by the natural gradient $\bar{F}^{-1}(\theta)\nabla \eta(\theta)$ and the projection. By Theorem 6.6.1 in \cite{Kushner2003}, the solution trajectory $\{\theta_n\}$ in Algorithm~1 also converges w.p.1 to the limit set of the ODE \eqref{eq:ODE}.
\hfill $\blacksquare$

\subsection{Approximation and Extension}\label{sec:extension}
In this section, we first discuss some approximations to make the proposed algorithm more practical. Note that in Algorithm~1, we use step-based natural policy gradient algorithm. It requires a single likelihood ratio per state-action pair. However, when computing the likelihood ratio, there is usually no closed-form expression for the discounted state visitation distribution $d^{\pi_{\theta}}(s)$. To make the algorithm more practical, we could replace the likelihood ratio $\omega(\xi,\theta_n|\theta_m)=\frac{d^{\pi_{\theta_n}}(\xi_m)}{d^{\pi_{\theta_m}}(\xi_m)}$ by $\widehat{\omega}(\xi,\theta_n|\theta_m)=\frac{\pi(\xi_m;\theta_n)}{\pi(\xi_m;\theta_m)}$ (e.g. \cite{degris2012off}). Recall that $\pi(\xi_m;\theta_n)=\pi_{\theta_n}(a_m|s_m)$, where $\xi_m=(s_m,a_m)$ is the state-action pair sampled at iteration $m$. Even though it introduces additional bias into the gradient estimator, we can show in the next corollary that the solution trajectory in Algorithm~1 with the likelihood ratio $\widehat{\omega}(\xi,\theta_n|\theta_m)$ converges w.p.1 to the same limit set of the ODE \eqref{eq:ODE}. 

\begin{corollary}\label{corollary:likelihood}
Under Assumption~\ref{ass:1}, the solution trajectory $\{\theta_n\}$ in Algorithm~1 with the likelihood ratio $\widehat{\omega}(\xi,\theta_n|\theta_m)$ converges w.p.1 to the limit set of the ODE \eqref{eq:ODE}.
\end{corollary}

It is natural to extend the proposed RNPG algorithm to some popular policy optimization algorithms such as TRPO. With a linear approximation to the objective and quadratic approximation to the constraint, the optimization in each iteration in TRPO can be written as
\begin{align*}
    & \max_{\theta} \quad \nabla \eta(\theta_n)(\theta - \theta_n)\\
    & \text{ s.t. } \quad \frac{1}{2}(\theta_n-\theta)^{T}F(\theta_n)(\theta_n-\theta) \leq \delta,
\end{align*}
where $F(\theta_n)_{ij}=\frac{\partial}{\partial \theta_i}\frac{\partial}{\partial \theta_j} \mathbb{E}_{s \sim d^{\pi_{\theta_n}}(s)}[D_{KL}(\pi_{\theta_n}(\cdot|s)\|\pi_{\theta}(\cdot|s))]|_{\theta=\theta_n}$ is the same FIM as in \eqref{eq:fisher}. $D_{KL}(P\|Q):=\int \log \left(\frac{d P}{d Q}\right) d P$ denotes the Kullback-Leibler divergence from distribution $P$ to distribution $Q$. 

Therefore, the update iterate in TRPO can be written as $\theta_{n+1}=\theta_n+\alpha_n F^{-1}(\theta_n) \nabla \eta(\theta_n)$. In practical implementation, TRPO performs a line search in the natural gradient direction, ensuring that the objective is improved while satisfying the nonlinear constraint. We can replace $F(\theta_n)$ and $\nabla \eta(\theta_n)$ by $\widehat{F}(\theta_n)$ and $\widehat{\nabla \eta}(\theta_n)$ in \eqref{eq:update_reuse} that reuse the historical trajectories while still ensuring the convergence of the TRPO algorithm, as guaranteed by Theorem~\ref{thm:ODE}.

\section{Characterization of Asymptotic Convergence Rate}\label{sec:rate}
In the section, we consider the asymptotic properties of normalized errors about the limit point obtained by RNPG and show that with diminishing step size, reusing historical trajectories helps improve the variance asymptotically. In particular, the asymptotic convergence rate is characterized by the covariance of the limiting normal distribution of the error. In addition to the Assumption~\ref{ass:1}, we need one more set of assumption to characterize the convergence rate.
\begin{assumption}~
\begin{itemize}
    \item (A.2.1) $\exists \bar{\theta} \in \Theta$, such that $\lim_{n \to \infty} \theta_n = \bar{\theta}$ w.p.1.
    \item (A.2.2) Let $\Sigma_{\eta}(\theta) = \operatorname{Var}(G(\xi,\theta))$ and recall $\Bar{F}^{-1}(\theta) = \mathbb{E}\left[\left(\addeps \frac{1}{B}\sum_{i=1}^B S(\xi_i,\theta) \right)^{-1} \right]$, where $\xi_1,\ldots,\xi_B$ are $i.i.d.$ samples from the occupancy measure $d^{\pi_{\theta}}$. Both $\Sigma_\eta(\theta)$ and $\Bar{F}^{-1}(\theta)$ are continuous in $\theta$. 
    \item (A.2.3) There exists a Hurwitz matrix $\mathcal{G}$  such that $\Bar{F}^{-1}(\theta) \nabla \eta (\theta) = \mathcal{G}(\theta - \bar{\theta}) + o(\|\theta-\bar{\theta}\|)$.
    \item (A.2.4) For $\theta \rightarrow \Bar{\theta}$, $S(\xi,\theta)$ converges to $S(\xi,\Bar{\theta})$ in distribution and $G(\xi,\theta)$ converges to $  G(\xi,\Bar{\theta})$ in distribution. 
    \item (A.2.5) The step size sequence $\{\alpha_n\}$ satisfies $\sum_{n=0}^{\infty} \alpha_{n}^{2}<\infty$, $\sum_{n=0}^{\infty} \alpha_{n}=\infty$, $\lim_{n \to \infty} \alpha_{n} = 0$, $\alpha_n > 0, \forall n \geq 0$, and $\sqrt{\frac{\alpha_n}{\alpha_{n+1}}} = 1 + o(\alpha_n)$.
\end{itemize}
\label{ass:2}
\end{assumption}

A sufficient condition for (A.2.1) is assuming a unique local maximum of $\eta(\theta)$ that lies in the interior of $\Theta$, as indicated by Theorem \ref{thm:ODE}.
(A.2.2) further implies $\Sigma_1(\theta):=\Bar{F}^{-1}(\theta) \Sigma_\eta (\theta) (\Bar{F}^{-1}(\theta))^{T}$ is continuous in $\theta$. We also define
\begin{align*}
    \Sigma_2'(\theta) := \mathbb{E}\left[\left(\addeps\frac{1}{B}\sum_{i=1}^B S(\xi_i,\theta) \right)^{-1} \Sigma_\eta(\theta) \left( \left(\addeps\frac{1}{B}\sum_{i=1}^B S(\xi_i,\theta) \right)^{-1} \right)^{T}   \right].
\end{align*}
(A.2.3) requires every eigenvalue of the matrix $\mathcal{G}$ has strictly negative real part, for the sake of asymptotic stability of the differential equation \eqref{eq:ODE}. This can be guaranteed if $\eta(\theta)$ is locally strongly concave in a neighborhood of $\Bar{\theta}$, where $\Bar{\theta}$ is the limiting point of the sequence $\{\theta_n\}$ in (A.2.1)  (please see  Appendix \ref{appsec: Hurwtiz} for the proof of this statement.
(A.2.5) covers a wide range of diminishing step size such as $\alpha_n = n^{-\beta}$ with $1/2<\beta<1$.


Next, let the normalized error  
\begin{align*}
    E_n=\frac{\theta_n-\bar{\theta}}{\sqrt{\alpha_n}},
\end{align*}
where sequence $\{\theta_n\}$ is obtained by the RNPG algorithm.  The normalized term $\sqrt{\alpha_n}$ represents the order of the convergence of $\theta_n$. By studying the convergence of the normalized error, we want to establish an exact characterization of the asymptotic convergence behavior of the error. 
Let $E_n(t)$ denote the piecewise constant right continuous interpolation of $\{E_m\}_{m\ge n}$. The following theorem shows the convergence of the normalized error process. 
\begin{theorem}\label{theorem:SDE}
Under Assumption~\ref{ass:1} and Assumption~\ref{ass:2}, there exists a Wiener process $W(t)$ with covariance matrix $\widehat{\Sigma}(\bar{\theta})$, such that $E_n(t)$ converges weakly to a stationary solution to the following SDE:
\begin{align*}
    d E = \mathcal{G} E dt + dW,
\end{align*}
where $\widehat{\Sigma}(\bar{\theta}) = \frac{1}{B}\Sigma_1(\Bar{\theta}) +\frac{1}{KB} \Sigma_2(\Bar{\theta})$, and $\Sigma_2(\Bar{\theta}) = \Sigma'_2(\Bar{\theta})-\Sigma_1(\Bar{\theta})$.

As a result, 
$$E_n \Rightarrow \mathcal{N}(0,\Sigma_\infty) \text{ as } n\rightarrow \infty,$$ where $\Rightarrow$ means convergence in distribution and ${\Sigma_{\infty}}$ has the following expression: $$\operatorname{vec}(\Sigma_\infty)= -(\mathcal{G}\oplus\mathcal{G})^{-1} \operatorname{vec}\left(\widehat{\Sigma}(\Bar{\theta})\right),$$
where $\oplus$ is the Kronecker sum and vec is the stack operator.
\end{theorem}

For an $n \times n$ square matrix $A$ and an $m \times m$ square matrix $B$, $A \oplus B = A \otimes I_m + I_n \otimes B$, where $I_m$ and $I_n$ are identity matrices of size $m$ and $n$, respectively, and $\otimes$ is the Kronecker product, such that $A \otimes B = 
\begin{bmatrix}
  a_{11}B & \cdots & a_{1n}B\\
  \vdots & \ddots & \vdots \\
  a_{n1}B & \cdots & a_{nn}B
\end{bmatrix}$. The stack operator vec creates a column vector from a matrix $A$ by stacking the column vectors of $A=[a_1 a_2 \cdots a_n]$ below one another:
$\operatorname{vec}(A)=
\begin{bmatrix}
  a_{1} \\
  a_{2} \\
  \vdots \\
  a_{n} \\
\end{bmatrix}$. 

Theorem \ref{theorem:SDE} suggests that the random error given by the algorithm, \( \theta_n - \bar{\theta} \), approximately follows a multivariate normal distribution with mean \( 0 \) and covariance matrix \( \alpha_n \Sigma_\infty \) when $n$ is large. We refer the readers to Section~\ref{subsec: LQC} for the verification of the asymptotic normality on a linear quadratic control (LQC, see \cite{anderson2007optimal}) problem. With this approximation, we have,
\[
\theta_n -\Bar{\theta} \sim \mathcal{N}(0,\alpha_n \Sigma_\infty),
\]
where $\operatorname{vec}(\Sigma_\infty)= -(\mathcal{G}\oplus\mathcal{G})^{-1} \operatorname{vec}\left(\widehat{\Sigma}(\Bar{\theta})\right)$ and
$\widehat{\Sigma}(\bar{\theta}) = \frac{1}{B}\Sigma_1(\Bar{\theta}) +\frac{1}{KB} \Sigma_2(\Bar{\theta})$. 
In the expression of $\widehat{\Sigma}$, the first term $\Sigma_1$ characterizes the variation caused by the random samples for the gradient estimator, and the second term $\Sigma_2$ reflects the joint impact of random samples for both gradient and inverse FIM estimators. Reusing past samples in the previous $K-1$ iterations helps reduce the covariance caused by the joint impact of random samples for both gradient and inverse FIM by an order of $O(\frac{1}{K})$. 
However, as $K$ increases further, this benefit diminishes because the asymptotic variance $\widehat{\Sigma}$ becomes dominated by $\frac{1}{B}\Sigma_1(\Bar{\theta})$.  As a result, the convergence rate of $\theta - \Bar{\theta}$ can be characterized, when $n$ is large, as 
\( \theta - \Bar{\theta} = O(\alpha_n^{\frac{1}{2}} \sqrt{\frac{1}{K} + O(1)})\). Notably,
a larger $K$ also demands additional computational resources and memory to store historical samples. In practice, the reuse size $K$ can be chosen based on the computational budget constraints or by balancing convergence gains with computational efficiency and memory usage. We will further explore the impact of reuse size numerically in Section \ref{subsec:reuse_size}.


Before the formal proof of Theorem~\ref{theorem:SDE}, we first give a high-level proof outline. We decompose the estimator of the natural policy gradient as 
\begin{align*}
    \widehat{F}^{-1}(\theta_n) \widehat{\nabla \eta}(\theta_n) 
    = &\underbrace{\mathbb{E}_n[\widehat{F}^{-1}_n \widehat{\nabla \eta}_n]}_{g_n} + \underbrace{\widehat{F}^{-1}_n \widehat{\nabla \eta}_n - \mathbb{E}_n[\widehat{F}^{-1}_n]\mathbb{E}_n  [\widehat{\nabla \eta}_n]}_{\delta G_n}
    \\
    =& \underbrace{\mathbb{E}_n[\widehat{F}^{-1}_n]\mathbb{E}_n  [\widehat{\nabla \eta}_n]}_{g_n}+ \underbrace{(\widehat{F}^{-1}_n - \mathbb{E}_n[\widehat{F}^{-1}_n])\widehat{\nabla \eta }_n + \mathbb{E}_n [\widehat{F}_n^{-1}] (\widehat{\nabla \eta}_n - \mathbb{E}_n[\widehat{\nabla \eta}_n])}_{\delta G_n}.
\end{align*}

For ease of notations, denote $\widehat{F}_n^{-1}=\widehat{F}^{-1}(\theta_n)$, $\widehat{\nabla \eta}_n=\widehat{\nabla \eta}(\theta_n)$, $\mathbb{E}_n[\cdot]=\mathbb{E}[\cdot|\pmb{e}_n(\theta)=\pmb{e}_n,\theta]$. Note that $\{\delta G_n\}$ is a martingale difference sequence, which contains the noise introduced by the new samples, and $g_n$ is the conditional mean conditioned on the reused samples. The asymptotic variance of $\{E_n\}$ is determined by the asymptotic behavior of the estimator of the natural policy gradient, which can be further characterized through analysis of the two terms: the Markov noise $\delta G_n$ and the conditional mean $g_n$. We first characterize the asymptotic behavior of $\delta G_n$ in the following Lemma \ref{lem: noise 1}-\ref{lem: noise 3}.

\begin{lemma}\label{lem: noise 1}
Suppose Assumption~\ref{ass:1} and Assumption~\ref{ass:2} hold. Let 
\begin{align*}
    L_n = \mathbb{E}_n [\widehat{F}_n^{-1}] (\widehat{\nabla \eta} - \mathbb{E}_n[\widehat{\nabla \eta}_n]) = \mathbb{E}_n [\widehat{F}_n^{-1}]\frac{1}{KB}\sum_{i=1}^{B} (G(\xi_n^i, \theta_n) - \nabla \eta(\theta_n)).
\end{align*}

Then we have
\begin{align*}
    \lim_{n,m\rightarrow\infty} \frac{1}{m}\sum_{i=n}^{m+n-1} \mathbb{E}_n\left[ L_i L_i^{T}- \frac{1}{K^2 B}\Sigma_1(\Bar{\theta})\right] = 0 \quad  \text{w.p.1},
\end{align*}
where the limit is taken as $n$ and $m$ go to infinity simultaneously.
\end{lemma}

\begin{lemma} \label{lem: noise 2}
Suppose Assumption~\ref{ass:1} and Assumption~\ref{ass:2} hold. Let 
\begin{align*}
    R_n = (\widehat{F}^{-1}_n - \mathbb{E}_n[\widehat{F}^{-1}_n])\widehat{\nabla \eta }_n.
\end{align*}

Then we have
\begin{align*}
    \lim_{n,m \rightarrow \infty} \frac{1}{m}\sum_{i=n}^{n+m-1} \mathbb{E}_n \left[ R_i R_i^{T} - \frac{1}{KB}\Sigma_2(\Bar{\theta}) \right] = 0 \quad  \text{w.p.1},
\end{align*}
where $\Sigma_2(\theta) = \Sigma'_2(\theta)-\Sigma_1(\theta)$.
    The limit is taken as $n$ and $m$ go to infinity simultaneously.
\end{lemma}

\begin{lemma}\label{lem: noise 3}
Suppose Assumption~\ref{ass:1} and Assumption~\ref{ass:2} hold. We have
\begin{align*}
    \lim_{n,m\rightarrow \infty} \frac{1}{m} \sum_{i=n}^{n+m-1} \mathbb{E}_n \left[L_i R_i^{T}\right] = 0 \quad \text{w.p.1},
\end{align*}
where the limit is taken as $n$ and $m$ go to infinity simultaneously.
\end{lemma}

With Lemma~\ref{lem: noise 1}-\ref{lem: noise 3}, we have
\begin{align}\label{eq: martingale difference noise}
    \lim_{n,m \to \infty} \frac{1}{m} \sum_{i=n}^{m+n-1} \mathbb{E}_n\left[\delta G_i \delta G_i^{T}-\left(\frac{1}{K^2 B}\Sigma_1(\bar{\theta})+\frac{1}{K B} \Sigma_2(\bar{\theta})\right)\right] = 0 \quad \text{w.p.1},
\end{align}
where the limit is taken as $n$ and $m$ go to infinity simultaneously.

Next, we characterize the asymptotic behavior of $g_n$ with the fixed-state method (see, e.g. \cite{Kushner2003}). Denote
\begin{align*}
    g_i(\theta,\pmb{e}_i(\theta)) = \mathbb{E}\left[\widehat{F}^{-1}(\theta,\pmb{e}_i(\theta))\widehat{\nabla \eta}(\theta,\pmb{e}_i(\theta)) | \pmb{e}_i(\theta),\theta\right],  \quad i\geq n.
\end{align*}

\begin{lemma}\label{lem: A8.5}
Suppose Assumption~\ref{ass:1} and Assumption~\ref{ass:2} hold. We have
\begin{align}\label{eq: conditional mean}
    \lim_{n,m\rightarrow \infty} \frac{1}{m}\sum_{i=n}^{n+m-1} \mathbb{E} \left[ g_i(\Bar{\theta},\pmb{e}_i(\Bar{\theta})) g_i(\Bar{\theta}, \pmb{e}_i(\Bar{\theta}))^{T} - \frac{K-1}{K^2B}\Sigma_1(\Bar{\theta}) | \pmb{e}_n(\Bar{\theta})\right] = 0  \quad \text{w.p.1},
\end{align}
where the limit is taken as $n$ and $m$ go to infinity simultaneously.
\end{lemma}

Let$$\Gamma_n(\theta,\pmb{e}_n(\theta))=\sum_{i=n}^{\infty}\Pi(n, i)\mathbb{E}\left[g_i(\theta,\pmb{e}_i(\theta)) - \bar{F}^{-1}(\theta)\nabla \eta(\theta) | \pmb{e}_n(\theta)\right],$$ 
where $\Pi(n, i)=\prod_{j=n}^{i}(1-\alpha_j)$, and 
$$
\Lambda_n(\theta, \pmb{e}_n(\theta))=\mathbb{E}\left[\Gamma_{n+1}(\theta,e_{n+1}(\theta)) \left(\widehat{F}^{-1}(\theta,\pmb{e}_n(\theta)) \widehat{\nabla \eta}(\theta,\pmb{e}_n(\theta))\right)^{T} | \pmb{e}_n(\theta)\right].
$$ 

We then introduce the last lemma that characterizes the asymptotic joint behavior of the conditional mean $g_n$ and martingale noise $\delta G_n$.

\begin{lemma}\label{lem: A8.7}
Under Assumption~\ref{ass:1} and Assumption~\ref{ass:2}, we have
\begin{align}\label{eq: matrix valued function}
    \lim_{n,m \to \infty} \frac{1}{m} \sum_{i=n}^{n+m-1} \mathbb{E}\left[\Lambda_i(\Bar{\theta}, \pmb{e}_i(\Bar{\theta})) - \frac{K-1}{2KB}\Sigma_1(\bar{\theta}) 
    |\pmb{e}_n(\Bar{\theta})\right] = 0 \quad \text{w.p.1},
\end{align}
where the limit is taken as $n$ and $m$ go to infinity simultaneously.
\end{lemma}

With Lemma~\ref{lem: noise 1}-\ref{lem: A8.7}, we are now ready to prove Theorem~\ref{theorem:SDE}.

\textbf{Proof of Theorem~\ref{theorem:SDE}}
By Theorem 10.8.1 in \cite{Kushner2003}, there is a Wiener process $W(t)$ with covariance matrix 
\begin{align*}
    \widehat{\Sigma} &= \frac{1}{K^2 B}\Sigma_1(\bar{\theta}) + \frac{1}{K B}\Sigma_2(\bar{\theta}) + \frac{K-1}{K^2B} \Sigma_1(\bar{\theta}) + \frac{K-1}{2KB} \Sigma_1(\bar{\theta}) + \frac{K-1}{2KB} \Sigma_1^{T}(\bar{\theta})\\
    & = \frac{1}{B} \Sigma_1(\bar{\theta}) + \frac{1}{KB} \Sigma_2(\bar{\theta}),
\end{align*}
where $\frac{1}{K^2 B}\Sigma_1(\bar{\theta}) + \frac{1}{K B}\Sigma_2(\bar{\theta})$ is from \eqref{eq: martingale difference noise}, $\frac{K-1}{K^2 B} \Sigma_1(\bar{\theta})$ is from \eqref{eq: conditional mean}, and $\frac{K-1}{2KB} \Sigma_1(\bar{\theta})$, $\frac{K-1}{2KB} \Sigma_1^{T}(\bar{\theta})$ are from \eqref{eq: matrix valued function}. We then have $E_n(t)$ converges weakly to a stationary solution of 
    \begin{align*}
        d E = \mathcal{G} E dt + dW.
    \end{align*}
Furthermore, the above SDE is known as the multivariate Ornstein-Uhlenbeck process (e.g., see \cite{maller2009ornstein}) and the covariance matrix of the solution $E(t)$ converges to ${\Sigma_{\infty}}$ as $t\rightarrow \infty$, where 
$$\operatorname{vec}(\Sigma_\infty) =  -(\mathcal{G}\oplus\mathcal{G})^{-1} \operatorname{vec}\left(\widehat{\Sigma}(\Bar{\theta})\right).$$
\hfill $\blacksquare$

\begin{corollary}\label{corollary:convergence_rate}
Let $E_n'=\frac{\theta_n'-\bar{\theta}}{\sqrt{\alpha_n}}$, where sequence $\{\theta_n'\}$ is obtained by the VNPG algorithm. Let $E_n'(t)$ denote the piecewise constant right continuous interpolation of the $\{E_m'\}_{m\ge n}$. Then under Assumption~\ref{ass:1} and Assumption~\ref{ass:2}, there exists a Wiener process $W'(t)$ with covariance matrix $\widetilde{\Sigma}(\bar{\theta})$, such that $E_n'(t)$ converges weakly to a stationary solution to the following SDE:
\begin{align*}
    d E' = \mathcal{G} E' dt + dW',
\end{align*}
where $\widetilde{\Sigma}(\bar{\theta}) = \frac{1}{B}\Sigma_1(\Bar{\theta}) +\frac{1}{B} \Sigma_2(\Bar{\theta})$ is the covariance matrix of $W'(\cdot)$.
\end{corollary}

Corollary~\ref{corollary:convergence_rate} is proved in a similar manner as Theorem~\ref{theorem:SDE}. Note that the covariance of the normalized error in RNPG algorithm is reduced by an order of $O(\frac{1}{K})$ compared to the VNPG algorithm. Therefore, with diminishing step size, reusing historical trajectories in RNPG reduces the asymptotic variance and the estimation error compared to VNPG.

\section{Numerical Experiments}\label{sec:numerical}
In the numerical experiment, we demonstrate the performance improvement of RNPG over VNPG on cartpole and MuJoCo inverted pendulum, two classical reinforcement learning benchmark problems. Furthermore, we verify the asymptotic normality of the error in RNPG algorithm as shown in Theorem~\ref{theorem:SDE} on an LQC problem.

\subsection{Experiment Setting and Benchmarks}
In cartpole, the goal is to balance a pole on a cart by moving the cart left or right. The state is a four-dimensional vector representing position of the cart, velocity of the cart, angle of the pole, and velocity of the pole. The action space is binary: push the cart left or right with a fixed force. The environment caps episode lengths to 200 steps and ends the episode prematurely if the pole falls too far from the vertical or the cart translates too far from its origin. The agent receives a reward of one for each consecutive step before the termination. 

\begin{figure}[ht]
    \centering
    \includegraphics[width=0.3\textwidth]{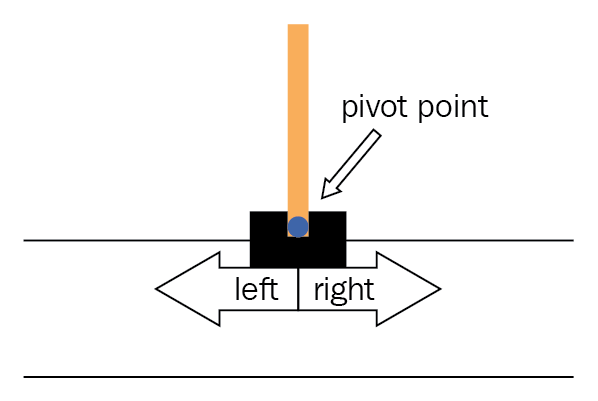}
    \caption{Diagram of cartpole and inverted pendulum task.}
    \label{fig:cartpole}
\end{figure}

MuJoCo stands for Multi-Joint dynamics with Contact. It is a physics engine for facilitating research and development in robotics, biomechanics, graphics and animation, and other areas where fast and accurate simulation is needed. We consider the inverted pendulum environment in MuJoCo. The inverted pendulum environment is the same as cartpole environment. The difference is that the action space is continuous in $[-3, 3]$, where action represents the numerical force applied to the cart (with magnitude representing the amount of force and sign representing the direction). The environment caps episode lengths to 500 steps. 

For both problems, the policy network is a fully-connected two-layer neural network with 32 neurons and Rectified Linear Unit (ReLU) activation function. We use softmax activation function on top of the neural network. The policy parameter is updated by Adam optimizer. We should note that similar performance can be obtained by using SGD optimizer with an appropriate learning rate. The discount factor is $\gamma=0.99$. We report the average reward over the number of iterations for different algorithms. The reward is averaged over $50$ macro replications. The number of trajectories generated in each iteration (i.e., mini-batch size) is $4$. We also set $\epsilon$ to 0.001 to prevent the FIM from becoming singular.
 
For both considered problems, we compare the performance of the following algorithms. 
\begin{itemize}
    \item VPG: vanilla policy gradient algorithm. 
    \item RPG: policy gradient algorithm with reusing historical trajectories.
    \item VNPG: vanilla natural policy gradient algorithm.
    \item RNPG: natural policy gradient algorithm with reusing historical trajectories.
\end{itemize}

\subsection{Experiment I: Convergence Rate on Cartpole and Inverted Pendulum}
\begin{figure}[ht]
    \centering
    \includegraphics[width=0.95\textwidth]{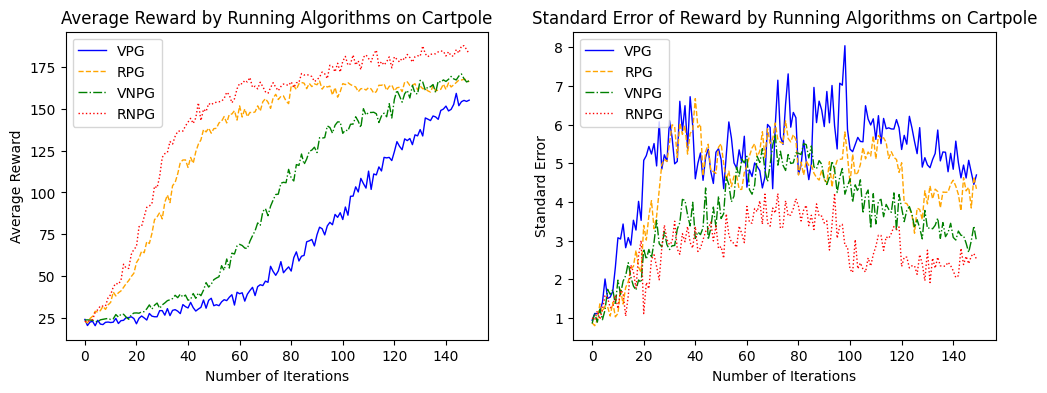}
    \caption{Mean (Figure 2(a)) and standard error (Figure 2(b)) of the reward over $n=150$ iterations for VPG, RPG, VNPG, and RNPG run on cartpole.}
    \label{fig:1}
\end{figure}

\begin{figure}[ht]
    \centering
    \includegraphics[width=0.95\textwidth]{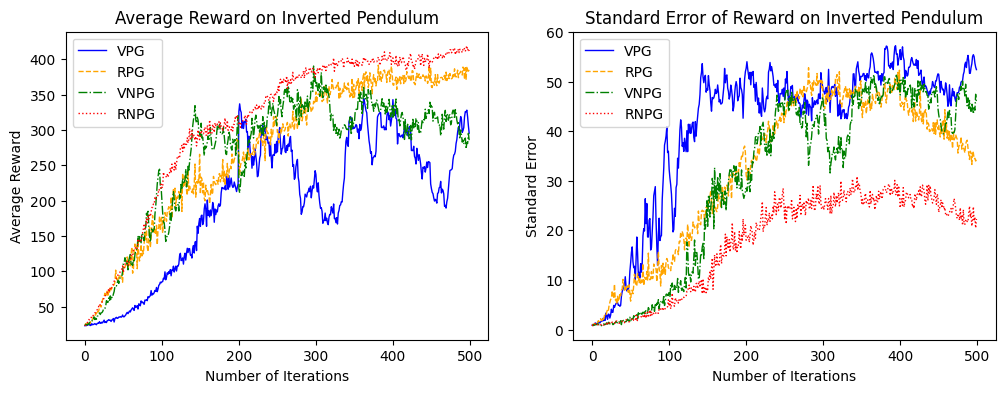}
    \caption{Mean (Figure 3(a)) and standard error (Figure 3(b)) of the reward over $n=500$ iterations for VPG, RPG, VNPG, and RNPG run on inverted pendulum.}
    \label{fig:2}
\end{figure}

In the first set of experiment, we run all aforementioned algorithms on cartpole and inverted pendulum problems, under a fixed step size $\alpha=0.01$ and the reuse size $K=10$. Specifically, the RNPG algorithm uses the same reuse size for both the FIM estimator and the gradient estimator. 
Figure~\ref{fig:1} and Figure~\ref{fig:2} show the mean and standard error of the reward for VPG, RPG, VNPG, and RNPG algorithms on cartpole benchmark over $150$ iterations and inverted pendulum benchmark over $500$ iterations, respectively. 
As can be seen from Figure~\ref{fig:1}(a) and Figure~\ref{fig:2}(a), reusing historical trajectories accelerates the convergence of the policy gradient algorithm (the convergence of RPG is faster than that of VPG) and the natural policy gradient algorithm (the convergence of RNPG is faster than that of VNPG). 
Both RPG and RNPG have a much smoother trajectory, compared with their vanilla counterpart VPG and VNPG. This can be seen from Figure~\ref{fig:1}(b) and Figure~\ref{fig:2}(b) the smaller standard errors of RPG and RNPG, compared to VPG and VNPG. It indicates that reusing historical trajectories reduces the variance of iterates and improves the stability of the algorithm. 

\begin{figure}[ht]
    \centering
    \includegraphics[width=0.95\textwidth]{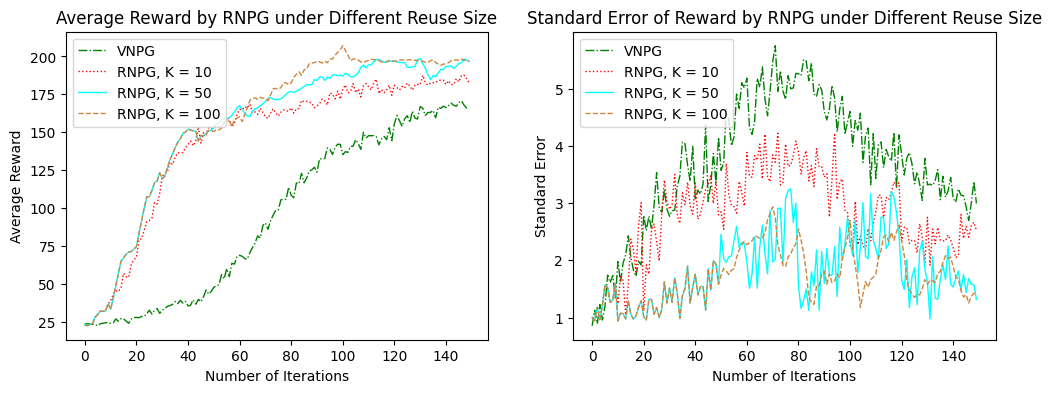}
    \caption{Mean (Figure 4(a)) and standard error (Figure 4(b)) of the reward over $n=150$ iterations for RNPG under reuse sizes $K=1,10,50,100$ run on cartpole.}
    \label{fig:3}
\end{figure}

\begin{figure}[ht]
    \centering
    \includegraphics[width=0.415\textwidth]{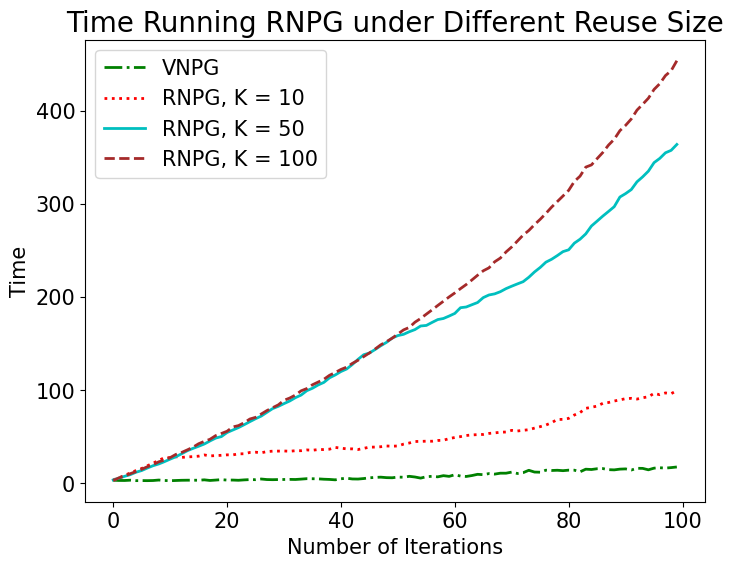}
    \includegraphics[width=0.42\textwidth]{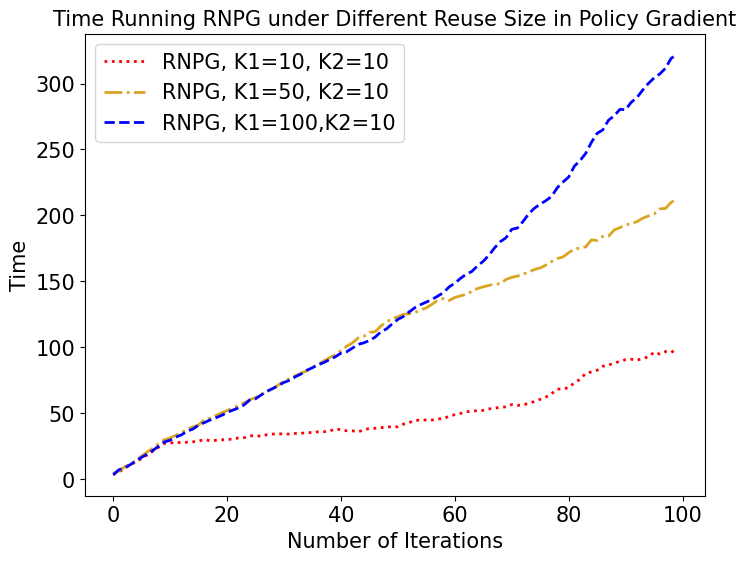}
    \caption{Time (s) running RNPG over $n=100$ iterations under different reuse sizes run on cartpole.}
    \label{fig:4}
\end{figure}

\begin{figure}[ht]
    \centering
    \includegraphics[width=0.95\textwidth]{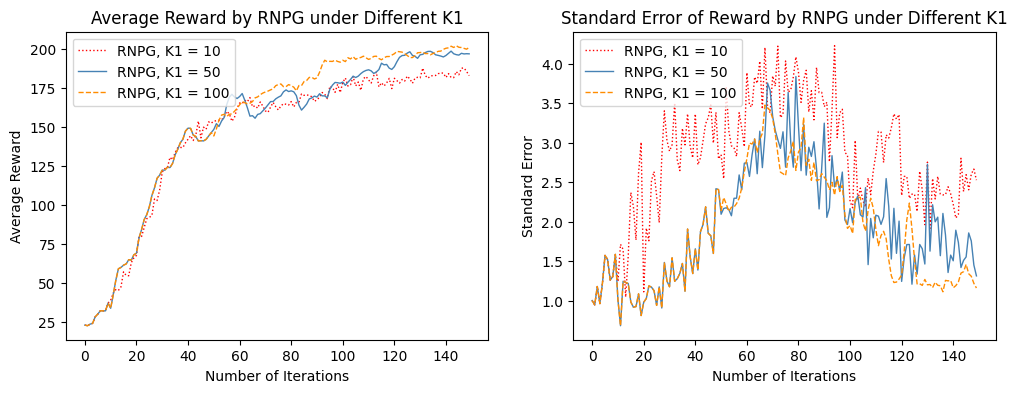}
    \caption{Mean (Figure 6(a)) and standard error (Figure 6(b)) of the reward over $n=150$ iterations for RNPG under reuse sizes $K_1=10,50,100$ run on cartpole. The reuse size $K_2$ is fixed to 10.}
    \label{fig:5}
\end{figure}

\subsection{Experiment II: Empirical Study on Reuse Size}\label{subsec:reuse_size}
In the second set of experiment, we empirically study the effect of the reuse size $K$ on the convergence rate of the RNPG algorithm. Specifically, the RNPG algorithm uses the same reuse size for both the FIM estimator and the gradient estimator. The step size is fixed to $\alpha=0.01$.
Figure~\ref{fig:3} shows the mean and standard error of the reward over $150$ iterations for RNPG algorithm with different reuse sizes $K=1, 10, 50, 100$ on the cartpole benchmark. Note that when $K=1$, RNPG is equivalent to VNPG, where we do not reuse any historical trajectory. 
As can be seen from Figure~\ref{fig:3}, when we reuse more historical trajectories from previous iterations (larger $K$), the faster the algorithm converges and the smoother the trajectory is. But this comes with the increased memory for computation. We report the average running time over $150$ iterations for RNPG algorithm with different reuse sizes on the cartpole benchmark in Figure~\ref{fig:4}(a). We should note that the main bottleneck is in computing the inverse FIM estimator with reusing historical trajectories. The computational complexity of computing the inverse FIM estimator in RNPG is $O(KBd^2+d^3)$, where $K$ is the reuse size in the FIM estimator. We further empirically study using different reuse sizes for the FIM estimator and the gradient estimator. 
In particular, Figure~\ref{fig:5} shows the mean and standard error of the reward over $150$ iterations for RNPG algorithm with reuse size $K_1=10, 50, 100$ in the gradient estimator and $K_2=10$ in the FIM estimator, on the cartpole benchmark problem. We also report the corresponding average running time in Figure~\ref{fig:4}(b). 
Figure~\ref{fig:5} suggests that we could use a reasonably small reuse size for the FIM estimator while using a large reuse size for the gradient estimator, such that we enjoy the benefit of reusing without sacrificing too much computational efficiency. 


\subsection{Reuse of Samples in Proximal Policy Optimization (RPPO)}

In previous sections, we evaluated the performance of policy gradient and natural policy gradient methods with and without reusing samples from past iterations. Here, we investigate the impact of sample reuse in Proximal Policy Optimization (PPO), a state-of-the-art policy optimization method. PPO updates the policy by maximizing the following surrogate function (see \cite{schulman2017proximal}):
$$
L^{C L I P}(\theta)=\frac{1}{B}\sum_{i=1}^B \min \left(r^i_n(\theta) {A}^i_n, \operatorname{clip}\left(r^i_n(\theta), 1-\epsilon, 1+\epsilon\right) {A}^i_n\right),
$$
where $r^i_n(\theta) = \frac{\pi_\theta(a_n^i|s_n^i)}{\pi_{\theta_n}(a_n^i|s_n^i)}$ is the likelihood ratio, $A_n^i$ is the advantage function, and $\operatorname{clip}(x,a,b)$ is the clipping function. 

To reuse past samples, we define the following surrogate function, incorporating samples from the previous $K-1$ iterations:
$$
L^{C L I P}_R(\theta)=\frac{1}{KB}\sum_{m=n-K+1}^n \sum_{i=1}^B \min \left(r^i_m(\theta) {A}^i_m, \operatorname{clip}\left(r^i_m(\theta), 1-\epsilon, 1+\epsilon\right) {A}^i_m\right).
$$
The updated solution $\theta_{n+1}$ maximizes $L^{C L I P}_R$. When $K=1$, $L^{C L I P}_R$ reduces to the original $L^{C L I P}$.

\begin{figure}[ht]
    \centering
\includegraphics[width=0.45\textwidth]{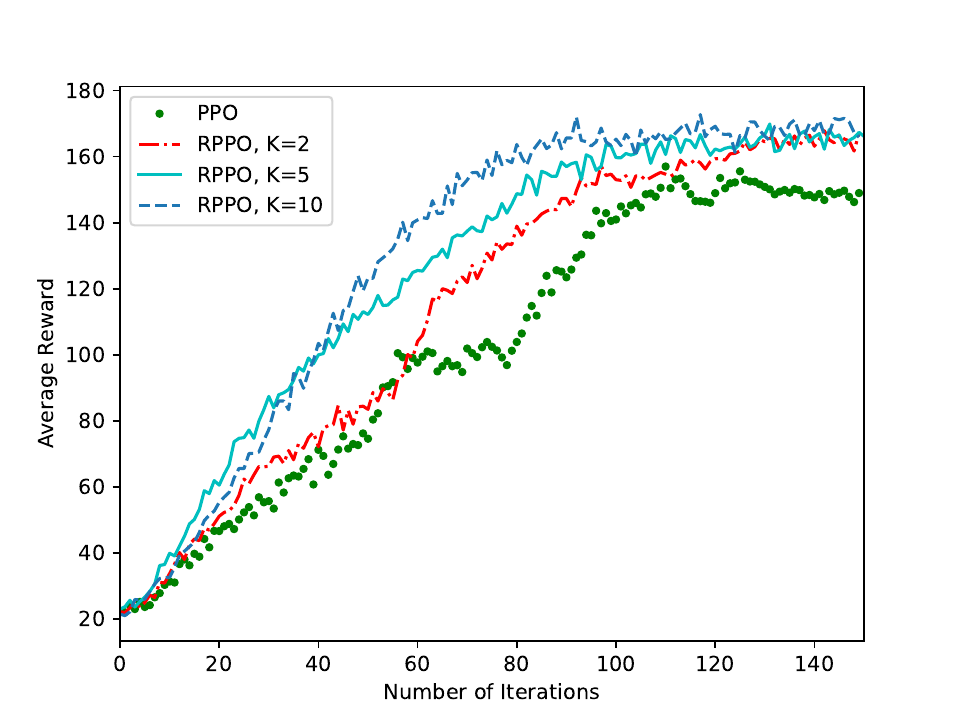}
\includegraphics[width = 0.45\textwidth]{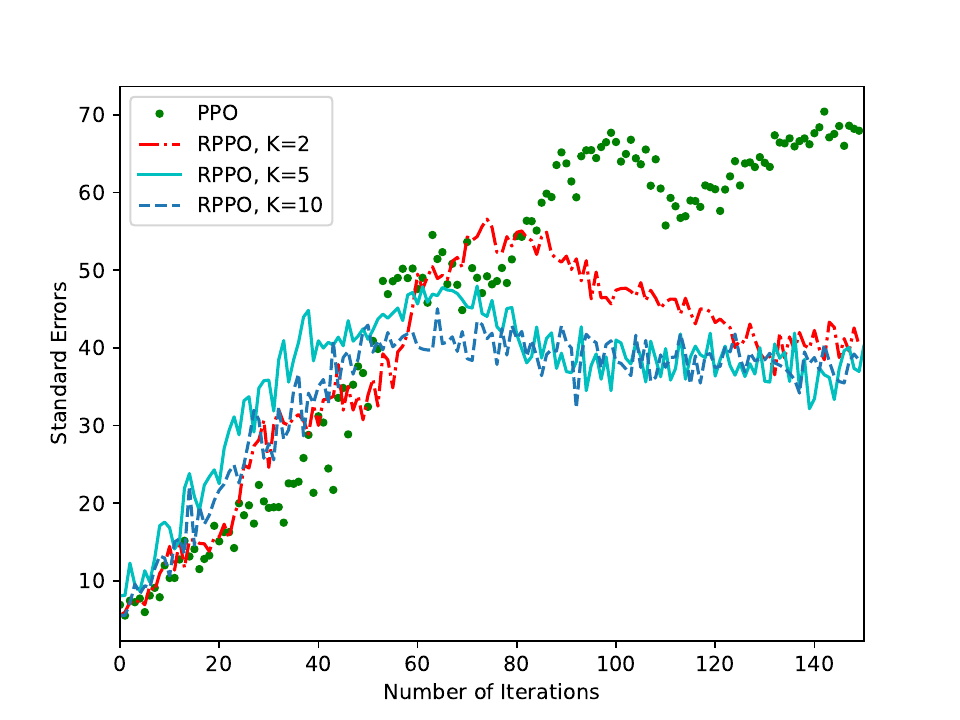}
    \caption{Mean (Figure \ref{fig: PPO}(a)) and standard error (Figure \ref{fig: PPO}(b)) of the reward over $n=150$ iterations for RPPO under reuse sizes $K=1,2,5$ run on cartpole.}
    \label{fig: PPO}
\end{figure}

We tested PPO and its sample-reuse variant (RPPO) on the same Cartpole environment introduced in \ref{sec:numerical}.1. The clip ratio $\epsilon$ is set to $0.2$. Figure \ref{fig: PPO} shows the mean and standard error of rewards over 150 iterations for both PPO and RPPO with reuse sizes $K=2,5$, averaged over 50 replications. Results indicate that RPPO achieves higher average rewards and lower standard errors with larger reuse sizes, demonstrating the benefit of sample reuse. 

\subsection{Verification of Asymptotic Normality on an LQC Problem}\label{subsec: LQC}
In this section, we numerically verify the asymptotic normality of the random error in the solution obtained by RNPG, as shown in Theorem \ref{theorem:SDE}. Consider the following one-dimensional Linear Quadratic Control (LQC) problem with discrete time and discounted cost:
\begin{equation} \label{eq: LQC}
    \begin{aligned}
    \min_{u=(u_1,u_2,\ldots)} \quad  & \mathbb{E}\left[\sum_{t=0}^\infty \gamma^t (x_t+u_t)^2\right]\\
    s.t. \quad & x_{t+1} = x_t + u_t + w_t,
\end{aligned}
\end{equation}
where $\gamma \in [0,1)$ is some discount factor, $x_t$ is the state at time $t$, $u_t$ is the action (or control) at time $t$, and $w_t \sim \mathcal{N}(0,1), t=0,1,\ldots$ are i.i.d. Gaussian noises. The expectation is taken with respect to all the randomness, which possibly contains the random initial state $x_0$ (which follows a standard normal distribution), random control $u_t(x_t), t \geq 0$, and the Gaussian noise $w_t,t\ge 0$. It is known that the optimal policy for \eqref{eq: LQC} is $u_t^*(x_t) = -x_t$ and the optimal objective is $\frac{1}{1-\gamma}$. To fit the LQC problem into the RL setting, we consider the discounted RL environment given by $\mathcal{M} = \left(\mathcal{S},\mathcal{A},\Theta, \mathcal{P},\mathcal{R},\gamma,\rho_0\right)$, where $\mathcal{S} = \mathbb{R}$ is the state space, $\mathcal{A} = \mathbb{R}$ is the action space, and $\Theta = \mathbb{R}$ is the policy parameter space. For $\theta \in \Theta, s\in \mathcal{S}$, $\pi_\theta(a|s) = \phi(a+s+\theta)$ and $\phi(\cdot)$ is the density function of the standard normal distribution; $\mathcal{P}$ is the transition kernel such that $\mathcal{P}_{s,a}(s') = 
\mathbf{1}_{\{s'=s+a\}}$, where $\mathbf{1}_{\{\cdot\}}$ is the indicator function; $\mathcal{R}$ is the reward function with $\mathcal{R}(s,a) = -(s+a)^2$; $\gamma = 0.5$ is the discount factor and $\rho_0$ is the initial distribution, which is the standard normal distribution. Here, $s$ corresponds to the state of the control problem $x$, and $a$ corresponds to the (disturbed) control $u+w$. The optimal policy $\pi^* = \pi_{\Bar{\theta}}$ with $\Bar{\theta} = 0$, and the optimal value function 
$\eta(\Bar{\theta}) = \mathbb{E}\left[\sum_{t=1}^\infty \gamma^t \mathcal{R}(s_t,a_t) \right] = -\frac{1}{1-\gamma}$. The optimal control $u^*_t$ can be recovered as $u^*_t = \Bar{\theta} - x_t = - x_t$. We refer the readers to Section~\ref{sec:LQC} in the appendix for the detailed calculation of the policy gradient, inverse FIM, and the theoretical asymptotic variance $\widehat{\Sigma}$ for the LQC problem.

\begin{figure}[h]
\begin{minipage}{.48\linewidth}
\centering
\includegraphics[width = \linewidth]{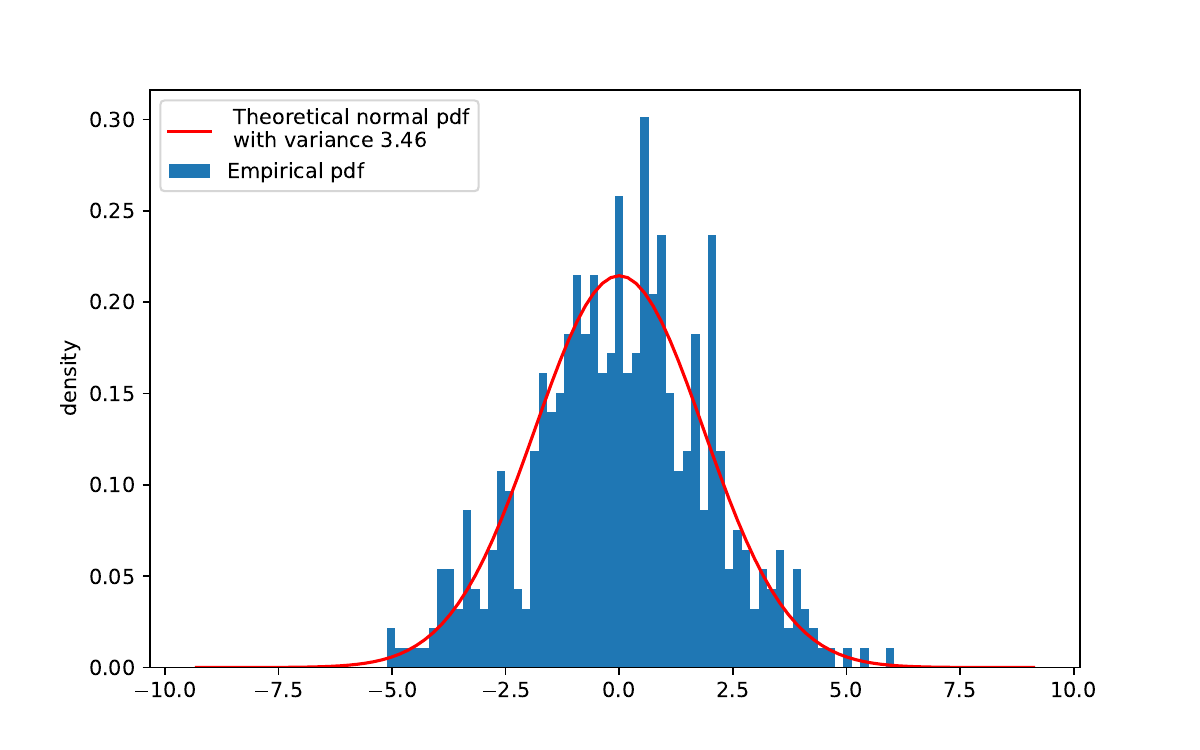}
{$B=5,K=1$}
\end{minipage}
~
\begin{minipage}{.48\linewidth}
\centering\includegraphics[width = \linewidth]{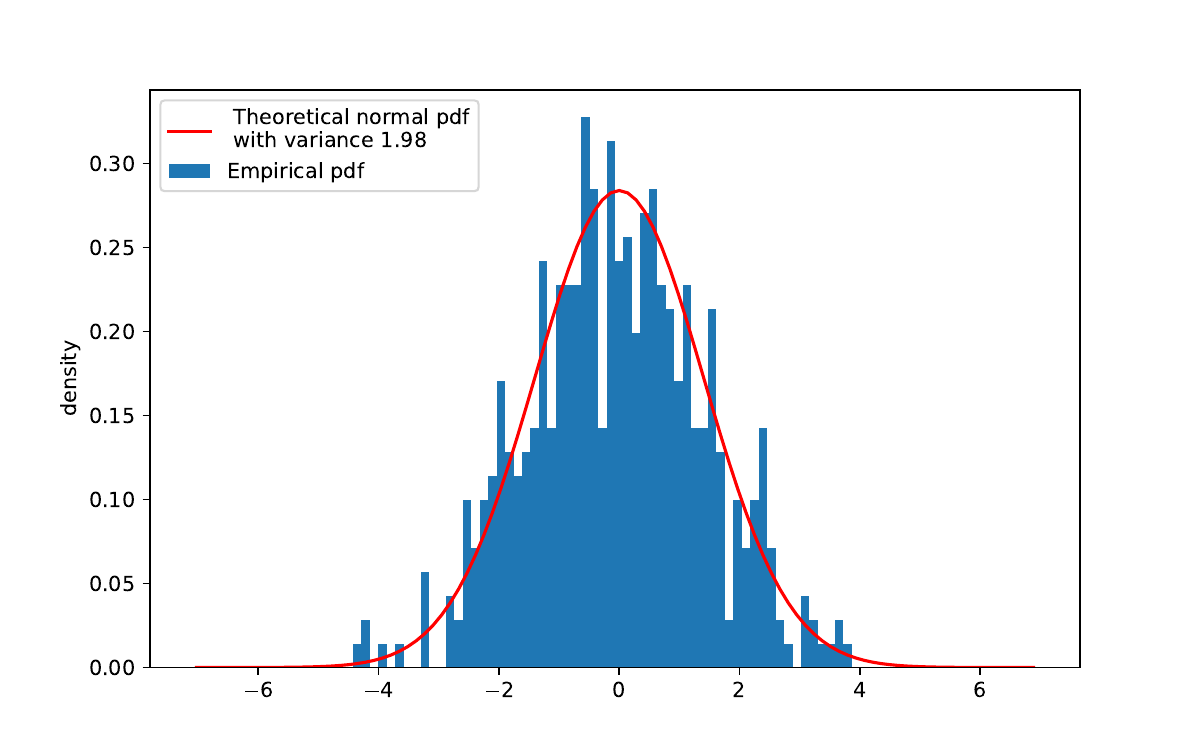}
{$B = 5, K=5$}
\end{minipage}

\begin{minipage}
{.48\linewidth}
\centering
\includegraphics[width = \linewidth]{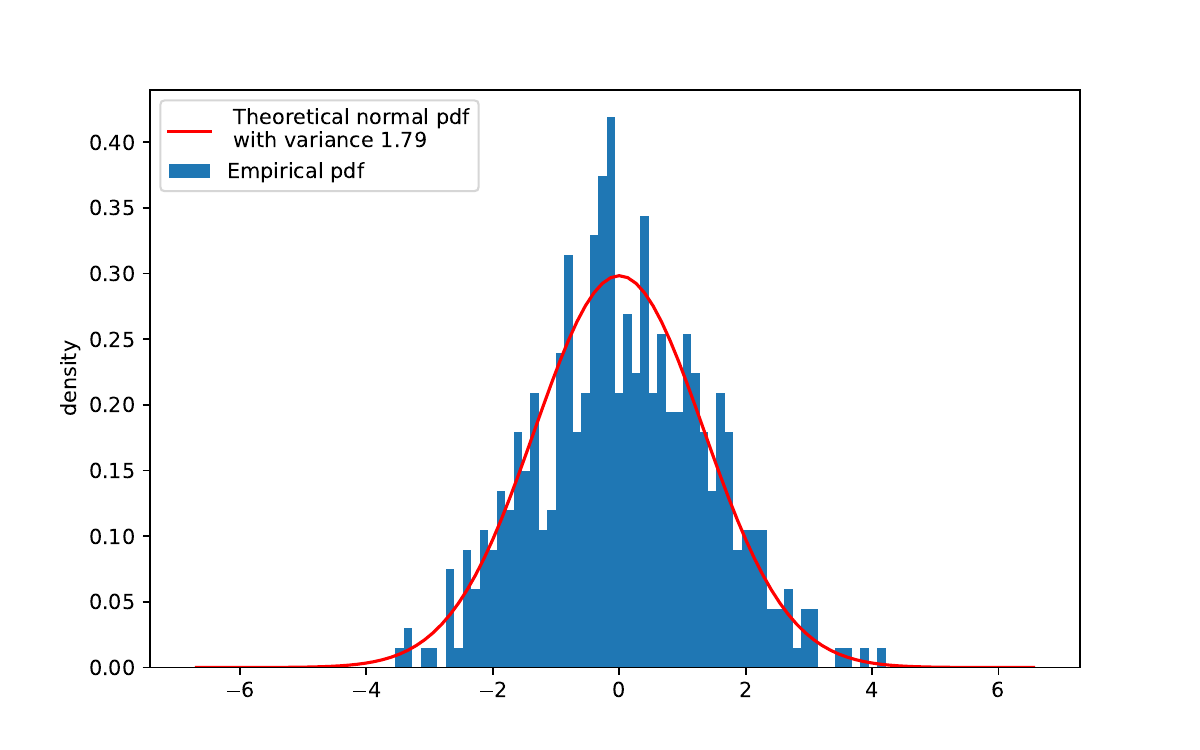}
{$B=5, K=10$}
\end{minipage}
~
\begin{minipage}{.48\linewidth}
\centering\includegraphics[width = \linewidth]{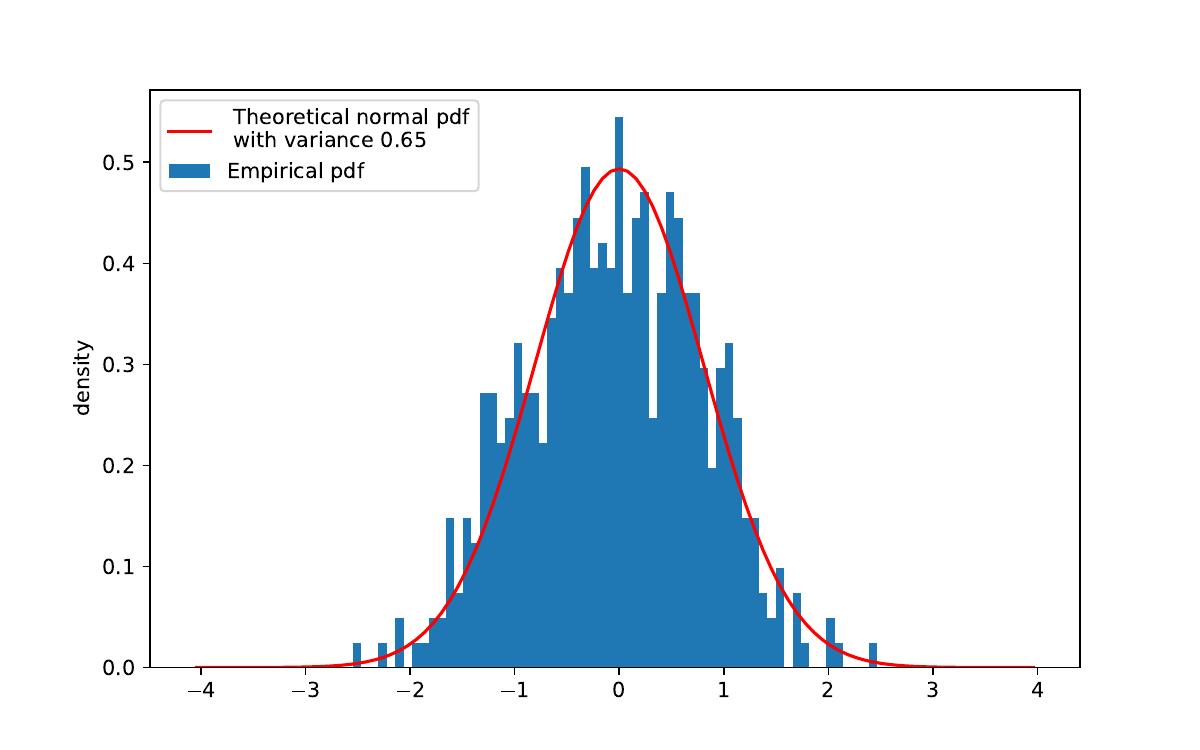}
{$B=10, K=5$}
\end{minipage}
\caption{Empirical density function and theoretical density function under different choices of $B$ and $K$.}
\label{fig: density}
\end{figure}

\begin{figure}[h]
\begin{minipage}{.48\linewidth}
\centering
\includegraphics[width = \linewidth]{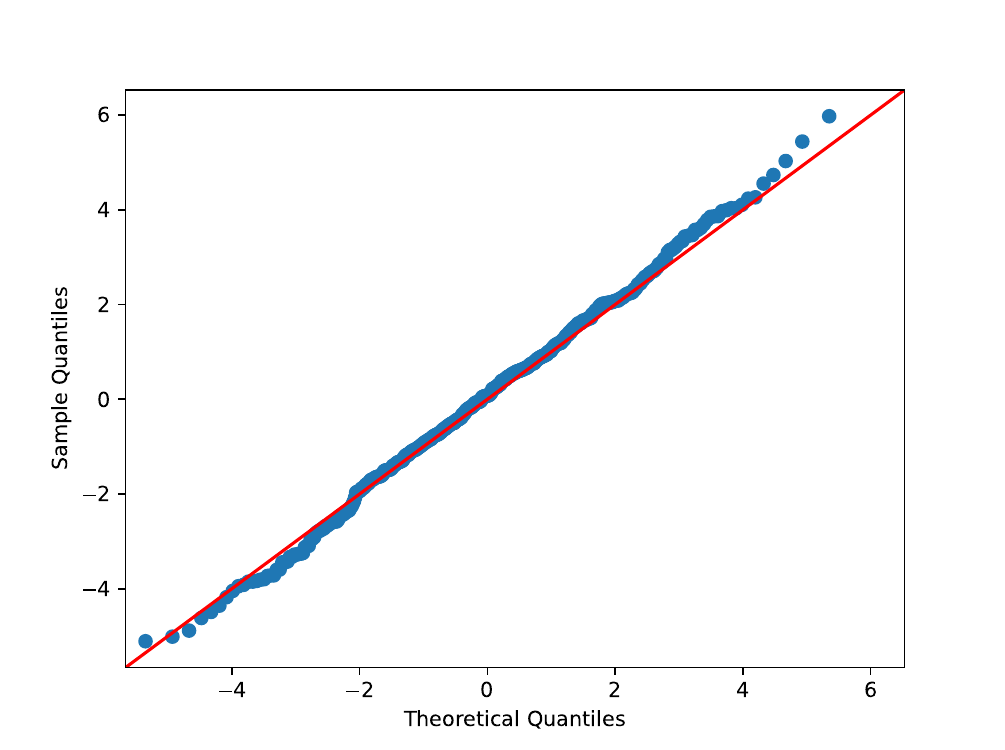}
{$B=5,K=1$}
\end{minipage}
~
\begin{minipage}{.48\linewidth}
\centering\includegraphics[width = \linewidth]{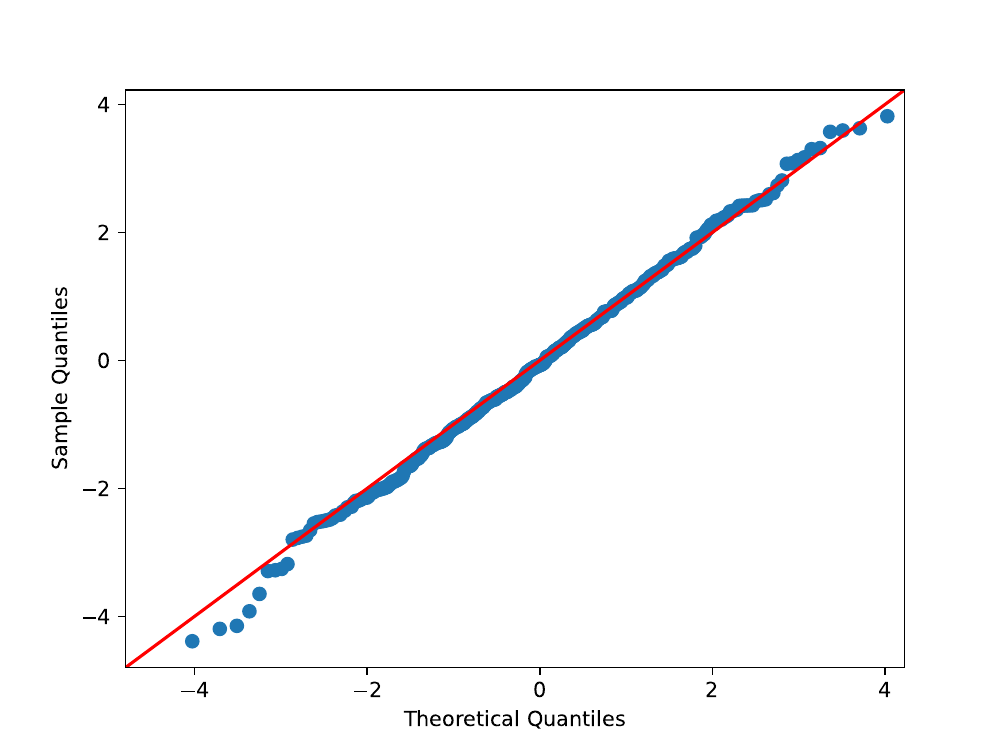}
{$B=5, K=5$}
\end{minipage}

\begin{minipage}{.48\linewidth}
\centering
\includegraphics[width = \linewidth]{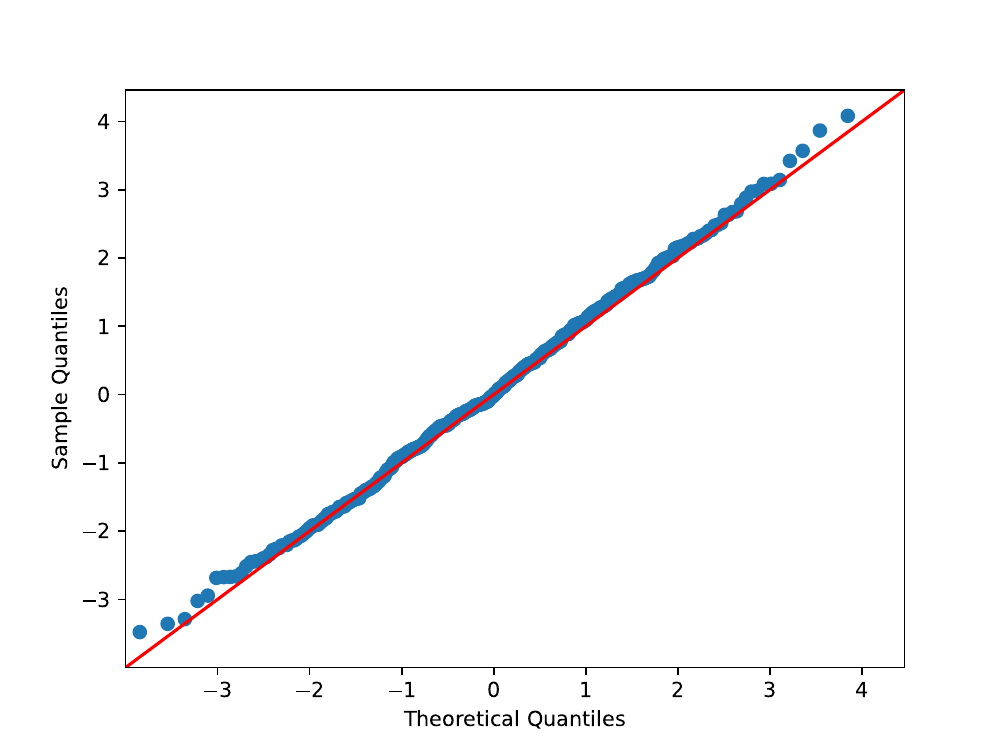}
{$B=5, K=10$}
\end{minipage}
~
\begin{minipage}{.48\linewidth}
\centering\includegraphics[width = \linewidth]{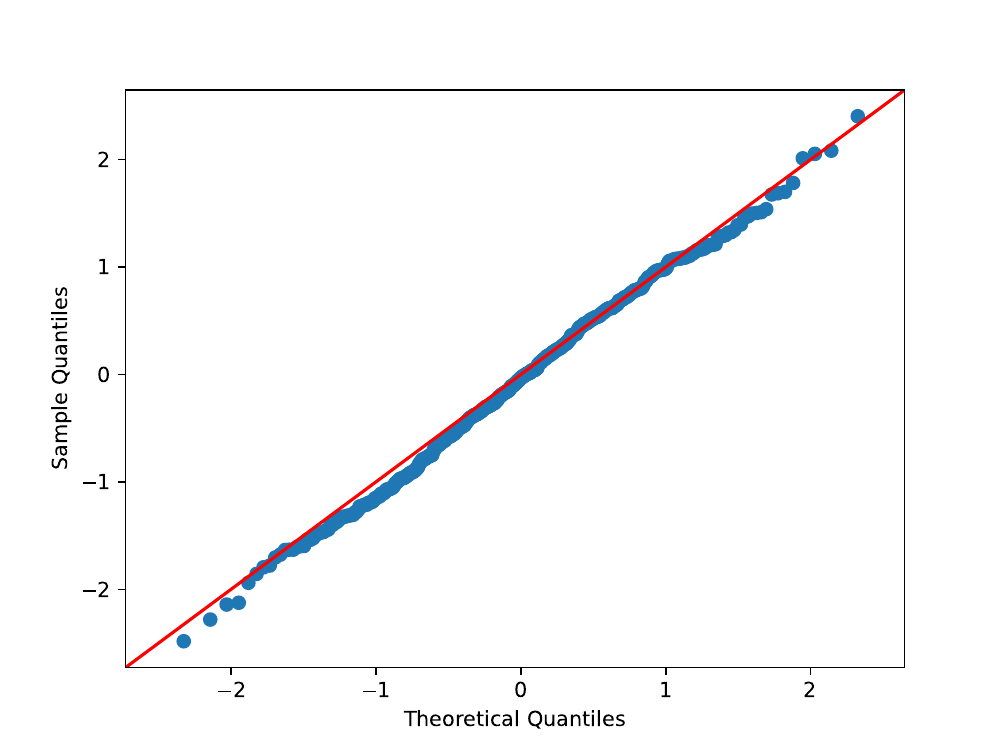}
{$B=10, K=5$}
\end{minipage}
\caption{Q-Q plot under different choices of $B$ and $K$.}
\label{fig: qq plot}
\end{figure}

To verify the asymptotic normality of $\frac{\theta_n-\Bar{\theta}}{\sqrt{\alpha_n}}$, we run $500$ macro-replications to plot the empirical density function and the quantile-quantile (Q-Q) plot. For each replication, we set $\alpha_n = 1/n^{0.9}$. The initial solution is set to $\theta_0 = 2$, and we run for $n = 5\times 10^5$ steps of the RNPG algorithm.
We vary the batch size $B$ and the reuse size $K$. In Figure \ref{fig: density}, we show the empirical density of $E_n$ and the theoretical density function for different choices of $B$ and $K$. In Figure \ref{fig: qq plot} we show the corresponding Q-Q plots. The high degree of overlap between the empirical density (quantile) function and theoretical density (quantile) function in Figure \ref{fig: density} (Figure \ref{fig: qq plot}) demonstrates the validity of Theorem \ref{theorem:SDE}. 

\section{Conclusion}\label{sec:conclusion}
In this paper, we study the convergence of a variant of natural policy gradient in reinforcement learning with reusing historical trajectories (RNPG). We provide a rigorous asymptotic convergence analysis of the RNPG algorithm by the ODE approach. Our results show that RNPG and its vanilla counterpart without reusing (VNPG) share the same limit ODE, while the bias resulting from the interdependence between iterations gradually diminishes, ultimately becoming insignificant in the asymptotic sense. We further demonstrate the benefit of reusing in RNPG and characterize the improved convergence rate of RNPG by the SDE approach. Through  numerical experiments on two classical benchmark problems, we demonstrate the performance of RNPG and  reusing samples in proximal policy optimization (PPO), and empirically study the choice of different reuse sizes. We also empirically verify the theoretical asymptotic normality result on a linear quadratic control problem. 

\section*{Acknowledgements}
 The authors are grateful for the support by Air Force Office of Scientific Research (AFOSR) under Grant FA9550-22-1-0244, National Science Foundation (NSF) under Grant ECCS-2419562, and Artificial Intelligence Institute for Advances in Optimization (AI4OPT) under Grant NSF-2112533. The first author (Yifan Lin) and second author (Yuhao Wang) contributed equally to the manuscript.

 \section*{Code and Data}
The code used for the experiments in Section \ref{sec:numerical} is available in the online supplementary materials.
 
\newpage
\appendix\section{Technical Proofs}\label{app:appendix}
Throughout the rest of the paper, for any vector $x \in \mathbb{R}^{d}$ or any matrix $X \in \mathbb{R}^{d \times d}$, let $\|\cdot\|$ denote the vector max norm (i.e., $\|x\|=\max \{|x_i|\}$) or the matrix max norm (i.e., $\|X\|=\max_{i,j}\{|x_{i,j}|\}$). Let $\|\cdot\|_2$ denote the vector 2-norm (i.e., $\|x\|_2=\sqrt{\sum_{i=1}^d x_i^2}$) or the matrix spectral norm (i.e., $\|X\|_2=\sqrt{\lambda_{\operatorname{max}}(X^{*} X)}$), where $X^{*}$ is the conjugate transpose of matrix $X$ and $\lambda_{\operatorname{max}}(\cdot)$ returns the largest eigenvalue. Let $\|\cdot\|_1$ denote the vector 1-norm (i.e., $\|x\|_1=\sum_{i=1}^{d} |x_i|$).

\subsection{Sufficient condition for (A.2.3)} \label{appsec: Hurwtiz}
\begin{lemma}
    Suppose $\eta(\theta)$ is locally strongly concave at $\Bar{\theta}$. Then (A.2.3) holds true. 
\end{lemma}

\textbf{Proof.}
Since $\eta(\theta)$ is locally strongly concave at $\Bar{\theta}$, there exists $\mu>0$, such that $\nabla^2 \eta(\theta) \preccurlyeq -\mu I_d$ for all $\theta$ in a small neighborhood of $\Bar{\theta}$. Then, we can write  
    $$\nabla \eta(\theta) = \nabla \eta(\Bar{\theta}) + \nabla^2\eta(\Bar{\theta})(\theta-\Bar{\theta}) + o(\|\theta-\Bar{\theta}\|) = \nabla^2\eta(\Bar{\theta})(\theta-\Bar{\theta}) + + o(\|\theta-\Bar{\theta}\|).$$
    Then
    \begin{align*}
        &\Bar{F}^{-1}(\theta) \nabla \eta(\theta) 
        \\= &\Bar{F}^{-1}(\theta)\nabla^2\eta(\Bar{\theta})(\theta-\Bar{\theta}) + o(\|\theta-\Bar{\theta}\|)\\
        =&\Bar{F}^{-1}(\Bar{\theta})\nabla^2\eta(\Bar{\theta})(\theta-\Bar{\theta}) + (\Bar{F}^{-1}(\theta)-\Bar{F}^{-1}(\Bar{\theta}))\nabla^2\eta(\Bar{\theta})(\theta-\Bar{\theta}) + o(\|\theta-\Bar{\theta}\|)
    \end{align*}
    With regularity the assumption on $\pi(\theta)$ (Lipschitz continuity, A.1.3), we have (by Proof of Lemma 2) $(\Bar{F}^{-1}(\theta)-\Bar{F}^{-1}(\Bar{\theta})) = O(\|\theta-\Bar{\theta}\|)$. This leads to 
    $$ \Bar{F}^{-1}(\theta) \nabla \eta(\theta)  = \Bar{F}^{-1}(\Bar{\theta})\nabla^2\eta(\Bar{\theta})(\theta-\Bar{\theta}) + o(\|\theta-\Bar{\theta}\|).$$ Since
    $$\Bar{F}^{-1}(\Bar{\theta}) = \mathbb{E}\left[\left(\epsilon I_d + \frac{1}{B}\sum_{i=1}^B S(\xi_i,\Bar{\theta}) \right)^{-1} \right],$$
    $\Bar{F}^{-1}(\Bar{\theta})$ is positive definite. Furthermore, we know
    $\nabla^2 \eta(\Bar{\theta}) \preccurlyeq -\mu I_d$ negative definite. 
    To show $\Bar{F}^{-1}(\theta) \nabla \eta(\theta)$ is Hurwitz, it is sufficient to show $\Bar{F}^{-1}(\theta) \nabla \eta(\theta)$ has $d$ negative eigenvalues. Notably, $\Bar{F}^{-1}(\theta) \nabla \eta(\theta)$  is not necessarily negative definite since it may not be symmetric. In the following we show for any positive definite matrix $A$ and negative definite matrix $B$, $A,B \in \mathbb{R}^{d\times d}$, the multiplication $AB$ has $d$ negative eigenvalues.

    Since $A$ is positive definite, there exists positive definite matrix $P$, such that $P^2  = A$. Then,
    $$AB = P^2 B = P(PBP) P^{-1}.$$
    That is, $M := PBP$ is a similar matrix for $AB$, and $M$ has the same eigenvalues as $AB$.  Note $M^\top = P^\top B^\top P^\top = PBP = M$. $M$ is symmetric. Further notice $\forall x \in \mathbb{R}^d$ and $x\neq \mathbf{0}$, 
    $$x^\top M x = x^\top PBP x = x^\top P^\top BP x =  (Px)^\top B (Px),$$
    since $P$ is positive definite, $Px \neq \mathbf{0}$. Furthermore, since $B$ is negative definite, $(Px)^\top B (Px) = x^\top M x < 0$. This implies $M$ is negative definite, and hence, $M$ has $d$ negative eigenvalues. This implies $AB$ also has $d$ negative eigenvalues, which means that $AB$ is a Hurwitz matrix.
\hfill $\blacksquare$

\subsection{Proof of Lemma~\ref{lemma:noise}}
\textbf{Proof.}
First note that $\mathcal{F}_n$ provides all the information required to achieve $\theta_n$, we have $\theta_n \in \mathcal{F}_n$. Moreover, conditioned on $\mathcal{F}_n$, the expectation of the gradient estimator with historical trajectories take the following form
\begin{align*}
    \bar{F}^{-1}(\theta_n) \mathbb{E}[\widehat{\nabla \eta}(\theta_n)|\mathcal{F}_n] = \frac{1}{KB} \sum_{m=n-K+1}^{n-1} \sum_{i=1}^{B} \omega(\xi_m^i,\theta_n|\theta_m) \bar{F}^{-1}(\theta_n)G(\xi_m^i,\theta_n) + \frac{1}{K} \bar{F}^{-1}(\theta_n) \nabla \eta(\theta_n).
\end{align*}
Then $\delta M_n=\bar{F}^{-1}(\theta_n) \widehat{\nabla \eta}(\theta_n)-\bar{F}^{-1}(\theta_n)\mathbb{E}[\widehat{\nabla \eta}(\theta_n)|\mathcal{F}_n]=\frac{1}{KB}\sum_{i=1}^{B}\bar{F}^{-1}(\theta_n)(G(\xi_n^i,\theta_n)-\nabla \eta(\theta_n))$. Note that (A.1.2) and (A.1.3) in Assumption~\ref{ass:1} together imply 
$$\|A^{\pi_{\theta}}(s,a) \nabla \pi_{\theta}(a|s)\|\leq \frac{2 U_r U_{\Theta}}{1-\gamma}, \forall (s,a) \in \mathcal{S} \times \mathcal{A}.$$
(A.1.2), (A.1.3) and (A.1.5) together imply the gradient $\nabla \eta(\theta)$ has bounded norm. It is then easy to check $\sup_n \mathbb{E}\|\delta M_n\|^2 < \infty$ from the boundedness assumption in Assumption~\ref{ass:1}. For large $j$ and some positive $T$, we have $\sum_{i=N(jT)}^{N(jT+t)-1} \alpha_i \leq 1$. Applying Burkholder's inequality (cf. Theorem 6.3.10 in \cite{stroock2010probability}), we have $\forall \epsilon > 0$, $\sum_{j} P\{\max_{0 \leq t \leq T} \|\sum_{i=N(jT)}^{N(jT+t-1)} \alpha_i \delta M_i\| \geq \epsilon\} < \infty$. Together with the step size $\sum_{n} \alpha_n^2 < \infty$ in Assumption~\ref{ass:1}, by Theorem 5.3.2 in \cite{Kushner2003}, $M(t)$ has zero asymptotic rate of change. 


For $H(t)$, note that we add a small perturbation $\epsilon I_d$ to the FIM to ensure its positive definiteness to prevent the FIM from becoming singular.  Hence, $\|\widehat{F}^{-1}(\theta_n)\| \leq \epsilon^{-1}$ almost surely. Also note that the norm of $\widehat{\nabla \eta}(\theta_n)$ is bounded w.p.1 under Assumption~\ref{ass:1}. Following the same argument in bounding the martingale difference sequence $\delta M_n$, we have $\sup_n \mathbb{E}\|\delta F_n\| < \infty$ and $\sup_n \mathbb{E}\|\delta F_n\|^2 < \infty$, and thus $H(t)$ has zero asymptotic rate of change.
\hfill $\blacksquare$

\subsection{Proof of Lemma~\ref{lemma:bias}}
\textbf{Proof.}
For ease of notations, denote $\mathbb{E}_n[\cdot]=\mathbb{E}[\cdot|\pmb{e}_n(\theta)=\pmb{e}_n,\theta]$. Note that $\forall \theta \in \Theta$, $\pmb{e}_i(\theta)$ is independent of $\pmb{e}_n(\theta)$ when $i \geq n + K -1$. So for those $i \geq n + K -1$, we have
\begin{align*}
    \mathbb{E}_n[\bar{F}^{-1}(\theta)(\widehat{\nabla \eta}(\theta, \pmb{e}_i(\theta)) - \nabla \eta(\theta))] = \frac{1}{KB}\sum_{s=i-K+1}^{i-1}\sum_{j=1}^{B} \mathbb{E}[\bar{F}^{-1}(\theta)G(\xi_s^j,\theta)\frac{d^{\pi_{\theta}}(\xi_s^j)}{d^{\pi_{\theta}}(\xi_s^j)}-\bar{F}^{-1}(\theta)\nabla \eta(\theta)]=0.
\end{align*}
We can then simplify $v_n(\theta_n,\pmb{e}_n)$ as $v_n(\theta_n,\pmb{e}_n)=\sum_{i=n}^{n+K-1}\alpha_i \mathbb{E}_n[\bar{F}^{-1}(\theta_n)\widehat{\nabla \eta}(\theta_n, \pmb{e}_i(\theta_n)) - \bar{F}^{-1}(\theta_n) \nabla \eta(\theta_n)]$. So
\begin{align*}
    \left\|\sum_{i=n}^{n+K-1} \alpha_i \mathbb{E}_n[\bar{F}^{-1}(\theta) (\widehat{\nabla \eta}(\theta, \pmb{e}_i(\theta))-\nabla \eta(\theta))]\right\| \leq \sum_{i=n}^{n+K-1} \alpha_i C_3 \to 0
\end{align*}
for some constant $C_3 > 0$. Here we use the inequality $\|M x\| \leq \|M\|\|x\|$, the fact that $\|M\| \leq \|M\|_2$ and $\|M\|_2$ is the largest eigenvalue for any positive definite matrix $M$ and vector $x$, under the condition (A.1.1), (A.1.2), (A.1.3) and (A.1.5) in Assumption~\ref{ass:1}. 
\hfill $\blacksquare$

\subsection{Proof of Lemma~\ref{lemma:fixed_chain}}
\textbf{Proof.}
Recall that for $m \leq n+K-2$, we have 
\begin{align*}
    \mathbb{E}_n[\bar{F}^{-1}(\theta) (\widehat{\nabla \eta}(\theta, \pmb{e}_m(\theta)) - \nabla \eta(\theta))] = \frac{1}{KB}\sum_{s=m-K+1}^{n-1}\sum_{i=1}^{B}\bar{F}^{-1}(\theta) (G(\xi_s^i,\theta)\frac{d^{\pi_{\theta}}(\xi_s^i)}{d^{\pi_{\theta_s}}(\xi_s^i)} - \nabla \eta(\theta)).
\end{align*}
So $v_n(\theta,\pmb{e}_n)$ can be written as $v_n(\theta,\pmb{e}_n)=\sum_{m=n}^{n+K-2}\alpha_m \frac{1}{KB}\sum_{s=m-K+1}^{n-1}\sum_{i=1}^{B} \bar{F}^{-1}(\theta) (G(\xi_s^i,\theta) \frac{d^{\pi_{\theta}}(\xi_s^i)}{d^{\pi_{\theta_s}}(\xi_s^i)}-\nabla \eta(\theta))$. We show $v_n(\cdot,\pmb{e})$ is Lipschitz continuous uniformly in $n$ and $\pmb{e}$. Note that $\forall \theta_1, \theta_2, \theta' \in \Theta$, $\forall \xi=(s,a) \in \mathcal{S} \times \mathcal{A}$, 
\begin{align*}
    & \left\|G(\xi,\theta_1)\frac{d^{\pi_{\theta_1}}(\xi)}{d^{\pi_{\theta'}}(\xi)}-G(\xi,\theta_2)\frac{d^{\pi_{\theta_2}}(\xi)}{d^{\pi_{\theta'}}(\xi)}\right\| \\
    = & \left\|A^{\pi_{\theta_1}}(s,a) \nabla \log \pi_{\theta_1}(a|s) \frac{d^{\pi_{\theta_1}}(s) \pi_{\theta_1}(a|s)}{d^{\pi'}(\xi)} - A^{\pi_{\theta_2}}(s,a) \nabla \log \pi_{\theta_2}(a|s) \frac{d^{\pi_{\theta_2}}(s) \pi_{\theta_2}(a|s)}{d^{\pi'}(\xi)}\right\| \\
    = & \frac{1}{d^{\pi'}(\xi)} \left\|A^{\pi_{\theta_1}}(s,a) \nabla \pi_{\theta_1}(a|s) d^{\pi_{\theta_1}}(s)-A^{\pi_{\theta_2}}(s,a) \nabla \pi_{\theta_2}(a|s) d^{\pi_{\theta_2}}(s)\right\|.
\end{align*}
From Lemma 3.2 in \cite{zhang2020global}, under the condition (A.1.2), (A.1.3) in Assumption~\ref{ass:1}, we have
\begin{align*}
    \left\|A^{\pi_{\theta_1}}(s,a) \nabla \pi_{\theta_1}(a|s) - A^{\pi_{\theta_2}}(s,a) \nabla \pi_{\theta_2}(a|s)\right\| \leq L_1 \|\theta_1-\theta_2\|, \quad \text{for some } L_1 > 0.
\end{align*}
From Lemma 3 in \cite{achiam2017constrained}, under the condition (A.1.4) in Assumption~\ref{ass:1}, we have 
\begin{align*}
    \|d^{\pi_{\theta_1}} - d^{\pi_{\theta_2}}\| \leq \|d^{\pi_{\theta_1}} - d^{\pi_{\theta_2}}\|_1 \leq \frac{2\gamma}{1-\gamma} \mathbb{E}_{s \sim d^{\pi_{\theta_2}}(s)} [\|\pi_{\theta_1}(\cdot|s)-\pi_{\theta_2}(\cdot|s)\|_{TV}] \leq \frac{2\gamma}{1-\gamma} U_{\Pi} \|\theta_1-\theta_2\|.
\end{align*}
To show the FIM $F(\theta)$ is also Lipschitz in $\theta$, we first show the following intermediate result.

\begin{lemma}\label{lemma:Lipschitz}
Let $f(x): \mathbb{R}^d \to \mathbb{R}^d$ be Lipschitz continuous in $x$ with bounded norm. Then $M(x):=f(x) f(x)^{T}$ is also Lipschitz continuous in $x$. 
\end{lemma}
\textbf{Proof of Lemma~\ref{lemma:Lipschitz}}
Note that $M(x)-M(y)=f(x)(f(x)^{T} - f(y)^{T}) + (f(x)-f(y)) f(y)^{T}$. Suppose $\|f(x)\| \leq L_2$, $\|f(x)-f(y)\| \leq L_3 \|x-y\|$ for some $L_2, L_3 > 0$. Then we have $M(x)$ Lipschitz in $x$ as follows
\begin{align*}
    \|M(x)-M(y)\| & = \max_{i,j} |(M(x)-M(y))_{i,j}|\\
    & = \max_{i,j} \left|(f(x)(f(x)^{T}-f(y)^{T}) + (f(x)-f(y))f(y)^{T})_{i,j}\right|\\
    & \leq 2 L_2 L_3 \|x-y\|.
\end{align*}
\hfill $\blacksquare$

\noindent With Lemma~\ref{lemma:Lipschitz}, we have $\bar{F}(\theta)$ is Lipschitz continuous in $\theta$, thus $\bar{F}^{-1}(\theta) G(\xi;\theta)\frac{d^{\pi_{\theta}}(\xi)}{d^{\pi_{\theta'}}(\xi)}$ is Lipschitz continuous in $\theta$, uniformly in $\xi$ and $n$. In addition, from Lemma 3.2 in \cite{zhang2020global}, $\bar{F}^{-1}(\theta) \nabla \eta(\theta)$ is also Lipschitz continuous in $\theta$. Thus we prove that $v_n(\cdot,\pmb{e})$ is Lipschitz continuous uniformly in $n$ and $\pmb{e}$. One can then check that $b_n=O(\alpha_n^2)$, which implies the asymptotic rate of change of $B(t)$ is zero under the condition (A.1.1) in Assumption~\ref{ass:1}. Following the same argument in Lemma~\ref{lemma:noise}, we have $I(t)$ has zero asymptotic rate of change. 
\hfill $\blacksquare$

\subsection{Proof of Corollary~\ref{corollary:bias}}
\textbf{Proof.}
Relating the perturbed iteration to the original iteration, we have
\begin{align*}
    \theta_{n+1} = \theta_{n} + \alpha_n (\bar{F}^{-1}(\theta_n) \nabla \eta (\theta_n) + \delta M_n + \delta F_n + z_n) + b_n + \delta B_n + v_n(\theta_n, \pmb{e}_n) - v_{n+1}(\theta_{n+1}, \pmb{e}_{n+1}).
\end{align*}
Hence we have 
\begin{align*}
    \alpha_n \zeta_n = b_n + \delta B_n + v_n(\theta_n, \pmb{e}_n) - v_{n+1}(\theta_{n+1}, \pmb{e}_{n+1}).
\end{align*}
We can rewrite $Z(t)$ as follows:
\begin{align*}
    Z(t) & = \sum_{i=0}^{N(t)-1} b_i + \delta B_i + v_i(\theta_i, \pmb{e}_i) -  v_{i+1}(\theta_{i+1}, \pmb{e}_{i+1}) \\
    & = B(t) + I(t) + v_0(\theta_0, \pmb{e}_0) - v_{N(t)}(\theta_{N(t)}, \pmb{e}_{N(t)}).
\end{align*}
We then have $\|Z(jT+t) - Z(jT)\| \leq \|B(jT+t) - B(jT)\| + \|I(jT+t) - I(jT)\| + \|v_{N(jT+t)}(\theta_{N(jT+t)}) - \pmb{e}_{N(jT+t)}\| + \|v_{N(jT)}(\theta_{N(jT)}) - \pmb{e}_{N(jT)}\|$. By Lemma~\ref{lemma:bias} and Lemma~\ref{lemma:fixed_chain}, $Z(t)$ has zero asymptotic rate of change w.p.1. 
\hfill $\blacksquare$

\subsection{Proof of Corollary~\ref{corollary:likelihood}}
\textbf{Proof.}
For any $s > 0$, let $\pmb{\xi}_s:=(\xi_s^1,\cdots,\xi_s^{B})$, $\pmb{\pi}_s:=(\pi_{\theta}(\xi_s^1),\cdots,\pi_{\theta}(\xi_s^B))$, effective memory $\pmb{e}_s:=(\pmb{\xi}_{s-K+1}$, $\pmb{\pi}_{s-K+1}, \cdots, \pmb{\xi}_{s-1}, \pmb{\pi}_{s-1})$, and non-decreasing filtration $\mathcal{F}_n:=\sigma\{(\theta_s, \pmb{e}_s), s \leq n\}$. The rest of the proof follows by replacing the occupancy measure $d^{\pi_{\theta}}$ by the policy $\pi_{\theta}$. Note that there is a slight modification of (A.1.5) in Assumption~\ref{ass:1}: there exists a constant $0<\epsilon_{\pi}\leq 1$ such that the policy $\pi_{\theta}(s,a) \geq \epsilon_{\pi}, \forall (s,a) \in \mathcal{S} \times \mathcal{A}, \forall \theta \in \Theta$.  
\hfill $\blacksquare$

\subsection{Proof of Lemma~\ref{lem: noise 1}}
\textbf{Proof.}
Denote $\bar{F}^{-1}_n = \mathbb{E}_n[\widehat{F}^{-1}_n]$. For $i\ge n+K$, we have 
\begin{align*}
     \mathbb{E}_n\left[ L_i L_i^{T}\right] = &\frac{1}{K^2B^2} \mathbb{E}_n\left[ \Bar{F}^{-1}_i\left( \sum_{j=1}^{B}  \left(G(\xi_i^j, \theta_i) - \nabla \eta(\theta_i)\right) \right)
     \left( \sum_{j=1}^{B}  \left(G(\xi_i^j, \theta_i) - \nabla \eta(\theta_i)\right) \right)^{T} (\Bar{F}^{-1}_i)^{T}\right]\\
     =&\frac{1}{K^2B^2} \mathbb{E}_n\left[ \Bar{F}^{-1}_i \mathbb{E}_i\left[\left( \sum_{j=1}^{B}  \left(G(\xi_i^j, \theta_i) - \nabla \eta(\theta_i)\right) \right)
     \left( \sum_{j=1}^{B}  \left(G(\xi_i^j, \theta_i) - \nabla \eta(\theta_i)\right) \right)^{T} \right](\Bar{F}^{-1}_i)^{T}\right]\\
     =& \frac{1}{K^2B} \mathbb{E}_n\left[ \Bar{F}^{-1}_i \Sigma_\eta(\theta_i)(\Bar{F}^{-1}_i)^{T}\right].
\end{align*}
The second equality holds because of the tower property and that conditioned on $\theta_i$, $\Bar{F}^{-1}_i$ is a constant. Next, since $\theta_i \rightarrow \Bar{\theta}$ w.p.1, $\Bar{F}^{-1}(\theta)$ is continuous in $\theta$ and $\Theta$ is compact, by bounded convergence theorem we have 
$\mathbb{E}[L_iL_i^{T}] \rightarrow \frac{1}{KB^2} \Sigma_1 (\Bar{\theta})$ w.p.1. Since the first $K$ terms do not affect the average value as $m$ goes to infinity, we complete the proof. 
\hfill $\blacksquare$

\subsection{Proof of Lemma~\ref{lem: noise 2}}
\textbf{Proof.}
For any $i\ge n+K$, we have 
\begin{align*}
    \mathbb{E}_n \left[ R_i R_i^{T} \right] = \mathbb{E}_n \left[ \left(\widehat{F}^{-1}_i - \Bar{F}_i^{-1}\right)\widehat{\nabla \eta }_i \widehat{\nabla \eta }_i^{T} \left(\widehat{F}^{-1}_i - \Bar{F}_i^{-1}\right)^{T}\right].
\end{align*}
Notice that 
\begin{align*}
    \widehat{\nabla \eta }_i \widehat{\nabla \eta }_i^{T} & = \sum_{i_1,i_2 = i-K+1}^i\sum_{j_1,j_2 = 1}^B \omega( \xi_{i_1}^{j_1},\theta_{i_1}|\theta_i) \omega( \xi_{i_2}^{j_2},\theta_{i_2}|\theta_i)G(\xi_{i_1}^{j_1},\theta_{i_1}) G(\xi_{i_2}^{j_2},\theta_{i_2})^{T} \\
    & := \sum_{i_1,i_2 = i-K+1}^i\sum_{j_1,j_2 = 1}^B \omega_{i_1,j_1}\omega_{i_2,j_2} G_{i_1,j_1} G_{i_2,j_2}^{T}.
\end{align*}
By (A.2.1), $\theta_n \rightarrow \Bar{\theta}$ w.p.1., hence $\omega_{i_1,j_1},\omega_{i_2,j_2}cx  \rightarrow 1$ w.p.1  and can be ignored. Also by (A.2.4) and the (conditional) independent samples for $\widehat{F}^{-1}_i$ and $G_{i,j}$, for $(i_1,j_1) \neq (i_2,j_2)$, we have $(\widehat{F}^{-1}_i,G_{i_1,j_1},G_{i_2,j_2})| \theta_n,\pmb{e}_n$ converge weakly to $(\widehat{F}^{-1}(\{\xi_j\}_{j=1}^B,\Bar{\theta}),G(\xi_1',\Bar{\theta}),G(\xi_2',\Bar{\theta}))$ for almost every $\theta_n, \pmb{e}_n$, where $\xi_1,\ldots,\xi_B,\xi_1',\xi_2'$ are i.i.d. sample from $d^{\pi_{\Bar{\theta}}}$ and $ \widehat{F}^{-1}(\{\xi_j\}_{j=1}^B,\Bar{\theta}) = \left(\addeps \sum_{i=1}^B S(\xi_i,\Bar{\theta})\right)^{-1}$. Furthermore, since the stochastic gradient $G$ is bounded by (A.1.2) and (A.1.5) and $S(\xi,\theta) S(\xi,\theta)^{T}$ is $\epsilon$-positive definite by (A.1.5) and (A.1.3) for some $\epsilon$ that depends on $\epsilon_d$ and $U_\Theta$, we know $(\widehat{F}^{-1}_i - \Bar{F}_i^{-1}) G_{i_1,j_1} G_{i_2,j_2}' (\widehat{F}^{-1}_i - \Bar{F}_i^{-1})^{T}$ is uniformly bounded, hence uniformly integrable. This implies w.p.1,
\begin{align*}
   & \mathbb{E}_n \left[  \left(\widehat{F}^{-1}_i - \Bar{F}_i^{-1}\right) G_{i_1,j_1} G_{i_2,j_2}^{T} \left(\widehat{F}^{-1}_i - \Bar{F}_i^{-1}\right)^{T} \right]\\
   \rightarrow & \mathbb{E} \left[ \left(\widehat{F}^{-1}(\{\xi_j\}_{j=1}^B,\Bar{\theta}) - \Bar{F}^{-1}(\Bar{\theta}) \right)G(\xi_1',\Bar{\theta}) G(\xi_2',\Bar{\theta})^{T}  \left(\widehat{F}^{-1}(\{\xi_j\}_{j=1}^B,\Bar{\theta}) - \Bar{F}^{-1}(\Bar{\theta})\right)^{T}\right]\\
   =& \mathbb{E} \left[ \left(\widehat{F}^{-1}(\{\xi_j\}_{j=1}^B,\Bar{\theta}) - \Bar{F}^{-1}(\Bar{\theta}) \right)\mathbb{E} \left[ G(\xi_1',\Bar{\theta})\right] \mathbb{E} \left[  G(\xi_2',\Bar{\theta})\right]^{T}  \left(\widehat{F}^{-1}(\{\xi_j\}_{j=1}^B,\Bar{\theta}) - \Bar{F}^{-1}(\Bar{\theta})\right)^{T}\right]\\
   =& 0.
\end{align*}
Similarly, for $(i_1,j_1) = (i_2,j_2)$, $(\widehat{F}^{-1}_i,G_{i_1,j_1})| \theta_n,\pmb{e}_n$ converges weakly to $(\widehat{F}^{-1}(\{\xi_j\}_{j=1}^B,\Bar{\theta}),G(\xi',\Bar{\theta}))$ for almost every $\theta_n, \pmb{e}_n$, where $\xi_1,\ldots,\xi_B,\xi'$ are i.i.d. samples from $d^{\pi_{\Bar{\theta}}}$. Hence, we have w.p.1.
\begin{align*}
   & \mathbb{E}_n \left[  \left(\widehat{F}^{-1}_i - \Bar{F}_i^{-1}\right) G_{i_1,j_1} G_{i_1,j_1}^{T} \left(\widehat{F}^{-1}_i - \Bar{F}_i^{-1}\right)^{T} \right]\\
   \rightarrow & \mathbb{E} \left[ \left(\widehat{F}^{-1}(\{\xi_j\}_{j=1}^B,\Bar{\theta}) - \Bar{F}^{-1}(\Bar{\theta}) \right)G(\xi',\Bar{\theta}) G(\xi',\Bar{\theta})^{T}  \left(\widehat{F}^{-1}(\{\xi_j\}_{j=1}^B,\Bar{\theta}) - \Bar{F}^{-1}(\Bar{\theta})\right)^{T}\right]\\
   =& \mathbb{E} \left[ \left(\widehat{F}^{-1}(\{\xi_j\}_{j=1}^B,\Bar{\theta}) - \Bar{F}^{-1}(\Bar{\theta}) \right)\mathbb{E} \left[ G(\xi',\Bar{\theta})  G(\xi',\Bar{\theta})^{T}\right]  \left(\widehat{F}^{-1}(\{\xi_j\}_{j=1}^B,\Bar{\theta}) - \Bar{F}^{-1}(\Bar{\theta})\right)^{T}\right]\\
   =& \mathbb{E} \left[ \widehat{F}^{-1}(\{\xi_j\}_{j=1}^B,\Bar{\theta} )\Sigma_\eta(\Bar{\theta}) \widehat{F}^{-1}(\{\xi_j\}_{j=1}^B,\Bar{\theta})^{T}\right] - \Bar{F}^{-1}(\Bar{\theta}) \Sigma_\eta(\Bar{\theta}) (\Bar{F}^{-1}(\Bar{\theta}))^{T}\\
   =& \Sigma'_2(\Bar{\theta}) - \Sigma_1(\Bar{\theta})\\
   =& \Sigma_2(\Bar{\theta}).
\end{align*}
Hence, we have 
$$ \mathbb{E}_n [R_i R_i^{T}] \rightarrow \frac{1}{KB} \Sigma_2(\Bar{\theta}) \quad \text{ w.p.1.}$$
Furthermore, since the first $K$ terms do not affect the average value when $m$ goes to infinity, we complete the proof. 
\hfill $\blacksquare$

\subsection{Proof of Lemma~\ref{lem: noise 3}}
\textbf{Proof.}
First, note that $\mathbb{E}_n\left[L_i R_i^{T}\right]=\mathbb{E}_n \left[ \left(\widehat{F}^{-1}_i - \mathbb{E}_i[\widehat{F}^{-1}_i]\right)\widehat{\nabla \eta }_i \left( \mathbb{E}_i [\widehat{F}_i^{-1}] \left(\widehat{\nabla \eta}_i - \mathbb{E}_i[\widehat{\nabla \eta}_i]\right)\right )^{T}\right]$. Since conditioned on $\theta_i,\pmb{e}_i$, $\widehat{F}_i^{-1}$ is independent of $\widehat{\nabla \eta}_i$, we can obtain
\begin{align*}
    & \mathbb{E}_n \left[\left(\widehat{F}^{-1}_i - \mathbb{E}_i[\widehat{F}^{-1}_i]\right)\widehat{\nabla \eta }_i \left( \mathbb{E}_i [\widehat{F}_i^{-1}] \left(\widehat{\nabla \eta}_i - \mathbb{E}_i[\widehat{\nabla \eta}_i]\right)\right )^{T}\right] \\
   = & \mathbb{E}_n \left[ \mathbb{E}_i \left[ \left(\widehat{F}^{-1}_i - \mathbb{E}_i[\widehat{F}^{-1}_i]\right)\right] \mathbb{E}_i\left[\widehat{\nabla \eta }_i \left( \mathbb{E}_i [\widehat{F}_i^{-1}] \left(\widehat{\nabla \eta}_i - \mathbb{E}_i[\widehat{\nabla \eta}_i]\right)\right)^{T}\right]\right] \\
    = & 0.
\end{align*}
Hence, the Lemma holds.
\hfill $\blacksquare$

\subsection{Proof of Lemma~\ref{lem: A8.5}}
\textbf{Proof.}
For $i \geq n+K$, we have
\begin{align*}
    g_i(\Bar{\theta},\pmb{e}_i(\theta)) & = \frac{1}{KB}\Bar{F}^{-1}(\Bar{\theta}) \left[ \sum_{j=i-K+1}^{i-1}\sum_{\ell=1}^B G(\xi_{j}^\ell,\Bar{\theta}) + B\nabla \eta(\Bar{\theta}) \right]  \\
    & = \frac{1}{KB}\Bar{F}^{-1}(\Bar{\theta}) \left[ \sum_{j=i-K+1}^{i-1}\sum_{\ell=1}^B G(\xi_{j}^\ell,\Bar{\theta})) \right],
\end{align*}
and $G(\xi_j^\ell)$ is independent of $\pmb{e}_n(\bar{\theta})$. Hence,
\begin{align*}
    &\mathbb{E} \left [ g_i(\Bar{\theta},\pmb{e}_i(\Bar{\theta})) g_i(\Bar{\theta}, \pmb{e}_i(\Bar{\theta}))^{T}| \pmb{e}_n(\Bar{\theta}),\Bar{\theta} \right] \\
    = & \frac{1}{K^2B^2} \Bar{F}^{-1}(\Bar{\theta}) \left(\sum_{i_1,i_2=i-K+1}^{i-1}\sum_{j_1,j_2=1}^B  \mathbb{E} \left[G(\xi_{i_1}^{j_1},\Bar{\theta}) G(\xi_{i_2}^{j_2},\Bar{\theta})^{T} \right]\right) (\Bar{F}^{-1}(\Bar{\theta}))^{T} \\
    =& \frac{(K-1)B}{K^2B^2} \Bar{F}^{-1}(\Bar{\theta})  \Sigma_\eta (\Bar{\theta}) (\Bar{F}^{-1}(\Bar{\theta}))^{T} \\
    =& \frac{K-1}{K^2 B} \Sigma_1(\Bar{\theta}).
\end{align*} 
Since the first $K$ terms does not affect the limit, we obtain the desired result.
\hfill $\blacksquare$

\subsection{Proof of Lemma~\ref{lem: A8.7}}
\textbf{Proof.}
Note that when $i \geq n + K$, $\pmb{e}_i(\theta)$ is independent of $\pmb{e}_n(\theta)$. So, we have 
\begin{align*}
    \mathbb{E}\left[g_i(\theta, \pmb{e}_i(\theta)) - \Bar{F}^{-1}(\theta)\nabla \eta(\theta) | \pmb{e}_n(\theta)\right]=0, i \geq n + K,
\end{align*}
and hence, we can write $\Gamma_n(\theta,\pmb{e}_n(\theta))$ as
\begin{align*}
\Gamma_n(\theta,\pmb{e}_n(\theta))=\sum_{i=n}^{n+K-1}\Pi(n, i)\mathbb{E}\left[g_i(\theta, \pmb{e}_i(\theta)) - \Bar{F}^{-1}(\theta) \nabla \eta(\theta) | \pmb{e}_n(\theta)\right].
\end{align*}

For $i\ge n+K $,  we have $\pmb{e}_i(\Bar{\theta})$ is independent of $\pmb{e}_n (\Bar{\theta})$. This implies 
\begin{align*}
    \mathbb{E}\left[ \Lambda_i(\Bar{\theta},\pmb{e}_i(\Bar{\theta})) | \pmb{e}_n(\Bar{\theta})\right] =  \mathbb{E}\left[ \Lambda_i(\Bar{\theta},\pmb{e}_i(\Bar{\theta})) \right].
\end{align*}
Since $\nabla \eta(\Bar{\theta}) = 0$, we can rewrite $\Gamma_{i+1}(\Bar{\theta},e_{i+1}(\Bar{\theta}))$ as
\begin{align*}
    \Gamma_{i+1}(\Bar{\theta},e_{i+1}(\Bar{\theta})) 
    =&\sum_{j=i+1}^{i+K-1} \Pi(i+1,j) \mathbb{E} \left[ g_j(\Bar{\theta},e_j(\Bar{\theta}) )|e_{i+1}(\Bar{\theta}) \right] \\
    = &\sum_{j=i+1}^{i+K-1} \Pi(i+1,j) \mathbb{E} \left[ \widehat{F}^{-1}(\Bar{\theta},e_j(\Bar{\theta})) \widehat{\nabla \eta} (\bar{\theta},e_j(\Bar{\theta}))|e_{i+1}(\Bar{\theta})\right] \\
    = & \sum_{j=i+1}^{i+K-1} \Pi(i+1,j) \Bar{F}^{-1}(\Bar{\theta})\mathbb{E} \left[  \widehat{\nabla \eta} (\bar{\theta},e_j(\Bar{\theta}))|e_{i+1}(\Bar{\theta})\right] \\
    = & \frac{1}{KB} \sum_{j=i+1}^{i+K-1} \Pi(i+1,j) \Bar{F}^{-1}(\Bar{\theta})  \sum_{\ell = j-K+1}^i \sum_{r=1}^B G(\xi_\ell^r,\Bar{\theta}).
\end{align*}
The third equality holds since conditioned on $e_{i+1}(\theta)$, $\widehat{F}_j^{-1}$ is independent of $\widehat{\nabla \eta}(\Bar{\theta},e_j(\Bar{\theta}))$. The fourth equality is obtained by writing $\widehat{\nabla \eta}(\Bar{\theta},\pmb{e}_i(\Bar{\theta})) = \frac{1}{KB}\sum_{\ell = i-K+1}^i \sum_{r=1}^B G(\xi_\ell^r,\Bar{\theta})$. Hence, we have 
\begin{align*}
    &\mathbb{E}\left[ \Lambda_i(\Bar{\theta},\pmb{e}_i(\Bar{\theta})) | \pmb{e}_n(\Bar{\theta})\right] \\
    = & \frac{1}{K^2B^2}  \sum_{j=i+1}^{i+K} \Pi(i+1,j) \Bar{F}^{-1}(\Bar{\theta})   \mathbb{E} \left[ \left( \sum_{\ell = j-K+1}^i \sum_{r=1}^B G(\xi_\ell^r,\Bar{\theta})\right) \left( \sum_{\ell = i-K+1}^i \sum_{r=1}^B G(\xi_\ell^r,\Bar{\theta})\right)^{T} \right] (\Bar{F}^{-1}(\bar{\theta}))^{T}\\
    =& \frac{1}{K^2B^2}  \sum_{j=i+1}^{i+K} \Pi(i+1,j)  (K+i-j) B\Bar{F}^{-1}(\Bar{\theta})  \Sigma_\eta (\Bar{\theta}) (\Bar{F}^{-1}(\bar{\theta}))^{T}.
\end{align*}
Since the step-size satisfies $\alpha_m\rightarrow 0$, we have $1 \ge \Pi(i+1,j) \ge \Pi(i+1,i+K) \ge \Pi(n+1,n+K) \rightarrow 1$ as $ n \rightarrow \infty$. Then, we have 
\begin{align*}
    \mathbb{E}\left[ \Lambda_i(\Bar{\theta},\pmb{e}_i(\Bar{\theta}))| \pmb{e}_n(\Bar{\theta})\right]  \rightarrow  \frac{K-1}{K B} \Sigma_1(\Bar{\theta}) \quad \text{w.p.1}.
\end{align*}
Finally, since the first $K$ ($i < n+ K$) terms do not affect the limit, we complete the proof. 
\hfill $\blacksquare$

\section{Calculation in the LQC Problem}\label{sec:LQC}
\subsection{Calculation of the Stochastic Gradient and the Inverse FIM Estimator}
Since $ \pi_\theta(a|s) = \phi(a+s+\theta)$, we have 
$\frac{\rd}{\rd_\theta} \pi_\theta(a|s) = \pi_\theta (a|s)(\theta - a - s)$, and hence $\frac{\rd}{\rd_\theta} \log \pi_\theta(a|s)  = \theta - a - s$.
Also, notice $s_t + a_t \sim \mathcal{N}(\theta,1)$ when $a_t \sim \pi_\theta(\cdot|s_t)$. We can compute the value function 
$V^{\pi_\theta}(s) = -\frac{1+\theta^2}{1-\gamma}$ and $Q^{\pi_\theta}(s,a) = -(s+a)^2 -\frac{\gamma }{1-\gamma} (1+\theta^2) = V^{\pi_\theta}(s) +1+\theta^2-(s+a)^2$, which gives us the advantage function $A^{\pi_\theta}(s,a) = 1+\theta^2 - (s+a)^2$. When $(s,a) \sim d^{\pi_\theta}$, $s+a \sim \mathcal{N}(\theta,1)$ as $a|s \sim \mathcal{N}(\theta-s,1)$. Hence, to get one sample of $G(\xi,\theta)$ and $S(\xi',\theta)$ for $\xi,\xi' \sim d^{\pi_\theta}$ and $\xi,\xi'$ independent, we can first sample $X,X'\sim\mathcal{N}(\theta,1)$, and compute $ G(\xi,\theta) = \frac{1}{1-\gamma}(1+\theta^2 - X^2)(\theta - X)$, $S(\xi',\theta) = (\theta - X)^2$. We also set the regularization term for inverse FIM $\epsilon = 0.01$. 

\subsection{Calculation of the Theoretical Asymptotic Variance}
To calculate the theoretical asymptotic $\widehat{\Sigma}$ in Theorem \ref{theorem:SDE}, we need to find the value of (i) $\Sigma_\eta(\Bar{\theta})$; (ii) $\Sigma_1(\Bar{\theta})$; (iii) $\Sigma_2'(\Bar{\theta})$; and (iv) $\mathcal{G}$. \\
For (i), we can compute
\begin{align*}
    \Sigma_\eta(\Bar{\theta}) = &\operatorname{Var} \left(G(\xi,\Bar{\theta})\right) \\
    = &\frac{1}{(1-\gamma)^2}\operatorname{Var}\left(A^{\pi_{\Bar{\theta}}} (s,a)  \frac{\rd}{\rd_{\Bar{\theta}}} \log\pi_{\Bar{\theta}}(a|s) \right)\\
    =& \frac{1}{(1-\gamma)^2}\operatorname{Var}\left( \left[(1-(s+a)^2\right]\left[ s+a\right]\right)\\
    =&\frac{1}{(1-\gamma)^2} \mathbb{E} \left[(s+a)^6 + (s+a)^2 - 2(s+a)^4 \right]\\
    =& \frac{10}{(1-\gamma)^2}, 
\end{align*}
where the variance and expectation is taken with respect to $(s,a) \sim d^{\pi_{\Bar{\theta}}}$, which leads to $s+a \sim \mathcal{N}(0,1)$.\\
For (ii), we have 
$$\Bar{F}^{-1} (\Bar{\theta}) = \mathbb{E}\left[ \left( \epsilon + \frac{1}{B} \sum_{i=1}^B S(\xi_i,\Bar{\theta})\right)^{-1}\right] = \mathbb{E}\left[ \left( \epsilon + \frac{1}{B} \sum_{i=1}^B X_i^2 \right)^{-1}\right],$$
where $X_1,X_2,\ldots,X_B$ are i.i.d. standard normal random variables. We compute the value of $\Bar{F}^{-1}(\Bar{\theta})$ by Monte Carlo simulation with $10^7$ replications. The value of $\Sigma_1(\Bar{\theta})$ can be obtained by $\Sigma_1(\bar{\theta})=\Bar{F}^{-2}(\bar{\theta}) \Sigma_\eta (\bar{\theta}) = \frac{10}{(1-\gamma)^2} \Bar{F}^{-2}(\Bar{\theta})$.\\
For (iii), we also use the Monte Carlo simulation with $10^7$ replications to compute the following expectation
\begin{align*}
    \Sigma_2'(\theta) = \Sigma_\eta(\theta)\mathbb{E}\left[\left(\epsilon+\frac{1}{B}\sum_{i=1}^B X_i \right)^{-2}   \right] = \frac{10}{(1-\gamma)^2} \mathbb{E}\left[\left(\epsilon+\frac{1}{B}\sum_{i=1}^B X_i \right)^{-2}   \right],
\end{align*}
where $X_1,X_2,\ldots,X_B$ are i.i.d. standard normal random variables. With (i)-(iii), we can then compute the value of $\widehat{\Sigma}$.\\
For (iv), we have 
\begin{align*}
    \frac{\rd}{\rd_\theta}\left( \Bar{F}^{-1}(\Bar{\theta}) \frac{\rd}{\rd_\theta} \eta(\Bar{\theta}) \right)
     =  \frac{\rd}{\rd_\theta} \Bar{F}^{-1} (\Bar{\theta}) \frac{\rd}{\rd_\theta} \eta(\Bar{\theta}) + \Bar{F}^{-1} (\Bar{\theta}) \frac{\rd^2}{\rd_{\theta^2}} \eta(\Bar{\theta}).
\end{align*}
Since $\eta(\theta) = -\frac{1+\theta^2}{1-\gamma}$, we can compute 
$\frac{\rd}{\rd_\theta} \eta(\theta) = -\frac{2\theta}{1-\gamma}$ and $ \frac{\rd^2}{\rd_{\theta^2}} \eta(\Bar{\theta}) = -\frac{2}{1-\gamma}$. For $\frac{\rd}{\rd_\theta} \Bar{F}^{-1} (\Bar{\theta}) $, we have 
\begin{align*}
    \frac{\rd}{\rd_\theta} \Bar{F}^{-1} (\theta)  = & \frac{\rd}{\rd_\theta} \mathbb{E}\left[\frac{1}{\epsilon + \frac{1}{B}\sum_{i=1}^B(X_i - \theta)^2 }\right]\\
    =& -\frac{2}{B}\mathbb{E}\left[\frac{\sum_{i=1}^B (X_i-\theta) }{\left(\epsilon + \frac{1}{B}\sum_{i=1}^B(X_i - \theta)^2 \right)^2}\right],
\end{align*}
where $X_1,\ldots,X_B$ are i.i.d. standard normal random variables. Then we have 
$$  \frac{\rd}{\rd_\theta} \Bar{F}^{-1} (\Bar{\theta})  =  -\frac{2}{B}\mathbb{E}\left[\frac{\sum_{i=1}^B X_i }{\left(\epsilon + \frac{1}{B}\sum_{i=1}^B X_i^2 \right)^2}\right] = 0,$$
where the last equality holds since the normal distribution is symmetrical. Then we know 
$$\mathcal{G} = -\frac{2}{1-\gamma} \Bar{F}^{-1}(\Bar{\theta}).$$
Together with $\mathcal{G}$ and $\widehat{\Sigma}(\Bar{\theta})$, we can then compute 
$\Sigma_\infty = -\frac{\widehat{\Sigma}(\Bar{\theta})}{2\mathcal{G}}$.

\newpage
\bibliographystyle{plain}
\bibliography{references}

\end{document}